\documentclass[lettersize,journal]{IEEEtran}
\usepackage{amsmath,amsfonts}
\usepackage{algorithmic}
\usepackage{algorithm}
\usepackage{array}
\usepackage[caption=false,font=normalsize,labelfont=sf,textfont=sf]{subfig}
\usepackage{textcomp}
\usepackage{stfloats}
\usepackage{float}
\usepackage{url}
\usepackage{verbatim}
\usepackage{graphicx}
\usepackage{cite}
\usepackage{colortbl}
\usepackage{makecell}
\usepackage{graphicx}
\usepackage{subfloat} 
\usepackage{mathrsfs}
\def\eg{\emph{e.g.}}

\def\ie{\emph{i.e.}}
\usepackage{graphicx}
\usepackage{multirow}
\usepackage{bbm}
\usepackage{hyperref}

\usepackage{color}

\begin{document}

\title{COMICS: End-to-end Bi-grained Contrastive Learning for Multi-face Forgery Detection}

\author{Cong Zhang, Honggang Qi, Shuhui Wang, Yuezun Li, ~\IEEEmembership{Member,~IEEE}, Siwei Lyu, ~\IEEEmembership{Fellow,~IEEE}
\thanks{Cong Zhang and Honggang Qi are with the University of Chinese Academy of Sciences, China. e-mail: (zhangcong171@mails.ucas.ac.cn;hgqi@ucas.ac.cn).}
\thanks{Shuhui Wang is with the Institute of Computing Technology, Chinese Academy of Sciences and also with Pengcheng Laboratory, Shenzhen 518066. e-mail:(wangshuhui@ict.ac.cn).}
\thanks{Yuezun Li is with the College of Computer Science and
Technology, Ocean University of China, China. e-mail: (liyuezun@ouc.edu.cn). }
\thanks{Siwei Lyu is with University at Buffalo, SUNY, USA. Email: (siweilyu@bufflao.edu).}
\thanks{Yuezun Li is {\em corresponding author}.}
\thanks{Copyright © 2024 IEEE. Personal use of this material is permitted. However, permission to use this material for any other purposes must be obtained from the IEEE by sending an email to pubs-permissions@ieee.org.}}
\markboth{Journal of \LaTeX\ Class Files,~Vol.~14, No.~8, August~2021}%
{Shell \MakeLowercase{\textit{et al.}}: A Sample Article Using IEEEtran.cls for IEEE Journals}


\maketitle

\begin{abstract} 
DeepFakes have raised serious societal concerns, leading to a great surge in detection-based forensics methods in recent years. Face forgery recognition is a standard detection method that usually follows a two-phase pipeline, {\it i.e.}, it extracts the face first and then determines its authenticity by classification. While those methods perform well in ideal experimental environment, they face challenges when dealing with DeepFakes in the wild involving complex background and multiple faces of varying sizes. Moreover, most face forgery recognition methods can only process one face at a time. One straightforward way to address this issue is to simultaneous process multi-face by integrating face extraction and forgery detection in an end-to-end fashion by adapting advanced object detection architectures. However, as these object detection architectures are designed to capture the discriminative features of different object categories rather than the subtle forgery traces among the faces, the direct adaptation suffers from limited representation ability. In this paper, we propose \underline{Co}ntrastive \underline{M}ulti-FaceForens\underline{ics} (COMICS), an end-to-end framework for multi-face forgery detection.  {COMICS integrates face extraction and forgery detection in a seamless manner and adapts to the advanced object detection architectures.} The core of the proposed framework is a bi-grained contrastive learning approach that explores face forgery traces at both the coarse- and fine-grained levels. Specifically, coarse-grained level contrastive learning captures the discriminative features among positive and negative proposal pairs at multiple layers produced by the proposal generator, and the fine-grained level contrastive learning captures the pixel-wise discrepancy between the forged and original areas of the same face and the pixel-wise content inconsistency among different faces. Extensive experiments on the OpenForensics and FFIW datasets demonstrate that our method outperforms other counterparts and shows great potential for being integrated into various architectures. Codes are available at \url{https://github.com/zhangconghhh/COMICS}.

\end{abstract}

\begin{IEEEkeywords}
DeepFake, Multi-face Forgery Detection, Contrastive Learning, Fine-grained Feature Learning.
\end{IEEEkeywords}

\section{Introduction}
\IEEEPARstart{D}{eepFake} refers to face forgery techniques using deep generative models \cite{li2019faceshifter,10141892}, which can synthesize highly realistic faces at scale, making it possible to create fake videos impersonating public persons or implanting faces of victims into pornographic videos conveniently. The abuse of DeepFake as a means of disinformation has raised serious public concerns and motivates the development of countermeasures against DeepFake \cite{wu2019mantra,yang2023masked}. 

\begin{figure}[t]
	\centering
	\includegraphics[width=\linewidth]{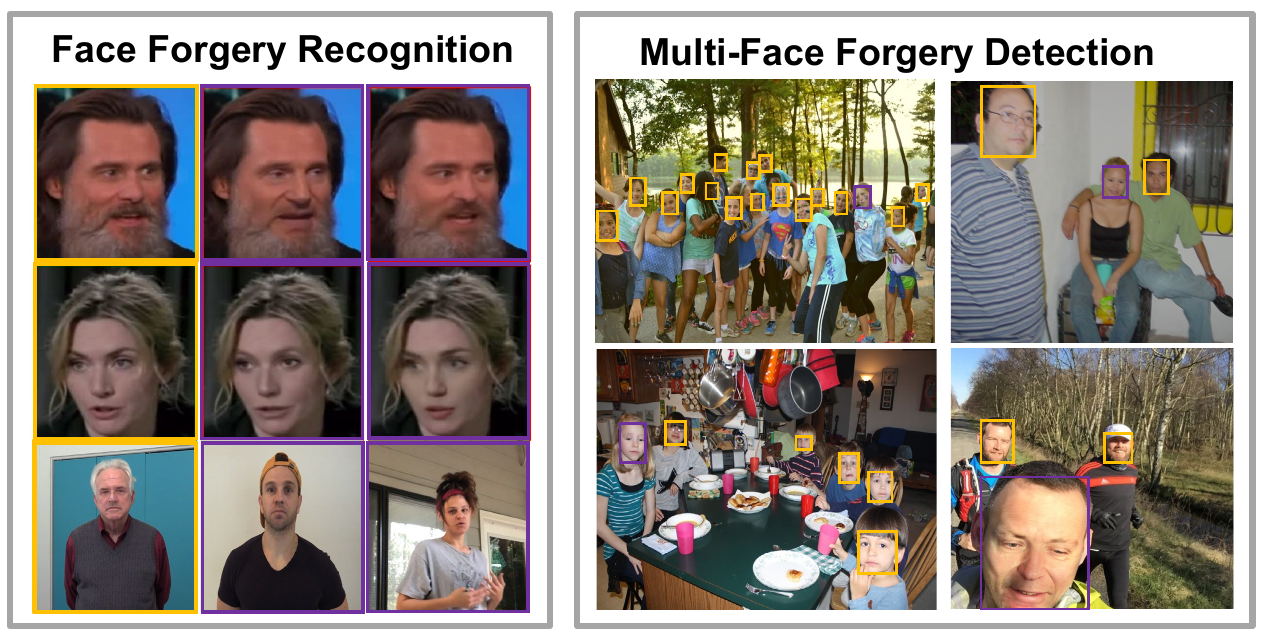} 
 	\vspace{-0.8cm}
    \caption{\small  Face forgery recognition usually deals with images of a fixed-size single face that has been carefully cropped. However, real-world images often contain complex backgrounds and multiple faces of varied sizes, making multi-face forgery detection more challenging. \textit{Images with yellow and purple boxes indicates the real faces and fake faces, respectively.}}    
    \vspace{-0.6cm}
	\label{fig:ourresult}
\end{figure}

\begin{figure*}[t]
	\centering
	\includegraphics[width=\linewidth]{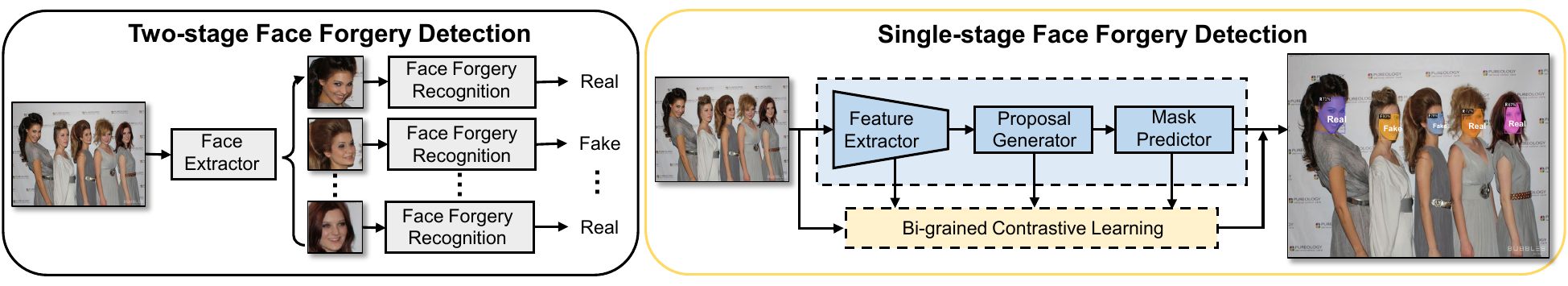} 
        \vspace{-0.7cm}
	\caption{\small Overview of the two-stage face forgery detection (left) and the proposed single-phase face forgery detection framework {\em Contrastive Multi-FaceForensics} (right).}
	\label{fig:twostage}
\end{figure*}

Currently, the majority of face forgery detection methods \cite{zhao2021multi,chen2021image,gu2022region} follow a two-stage pipeline known as {\em face forgery recognition} \cite{le2021openforensics}, {\it i.e.}, faces are first extracted using face detectors, and then they are classified as real or DeepFake (Fig.~\ref{fig:twostage} (left)). These methods perform well on images or videos that contain a single fixed-size face with a clear boundary against a simple background.
However, real-world images usually contain multiple faces with diverse rotation pattern, object sizes and complex backgrounds, as illustrated in Fig.~\ref{fig:ourresult}(b).
These two-stage methods are primarily designed to focus on the global discrepancy between real and fake faces, neglecting the relationship among multiple faces in one image and between faces and backgrounds.
Even worse, applying face forgery recognition in these scenarios appears to be problematic since they can only process face-by-face sequentially, resulting in longer running time or resource consumption.
Therefore, it is essential to develop a more efficient method capable of simultaneously processing multiple faces in an input image or video frame for real-world applications.

To facilitate the study to overcome the limitation of traditional face-to-face processing criteria, the OpenForenscis dataset \cite{le2021openforensics} was proposed to encourage the development of end-to-end pipelines that can simultaneously perform face detection and forgery detection\footnote{In this paper, face forgery segmentation is also considered. For simplicity, we use detection as a general term for detection and segmentation hereafter except for Sec.~\ref{sec:exps}.}. The straightforward solution is to adapt standard object detection architectures to this task directly \cite{lee2020centermask, zhang2020mask, tian2020conditional}. While the object detection based approaches have shown feasibility to identify multi-forgery faces, the object detectors are originally designed to learn discriminative features among object categories, rather than the subtle characteristics that distinguish real from DeepFake faces. Consequently, it does not yield satisfactory results (See Sec.~\ref{sec:exps-res}). Recent {studies} \cite{sun2022dual,chugh2020not} start to use contrastive learning to extract forged traces in face forgery recognition. However, these methods conduct contrastive learning between real and fake faces without considering the complex backgrounds and variations in faces. This limitation makes the direct application of contrastive learning less effective for multi-face forgery detection.

\begin{figure*}[t]
	\centering
	\includegraphics[width=0.9\linewidth]{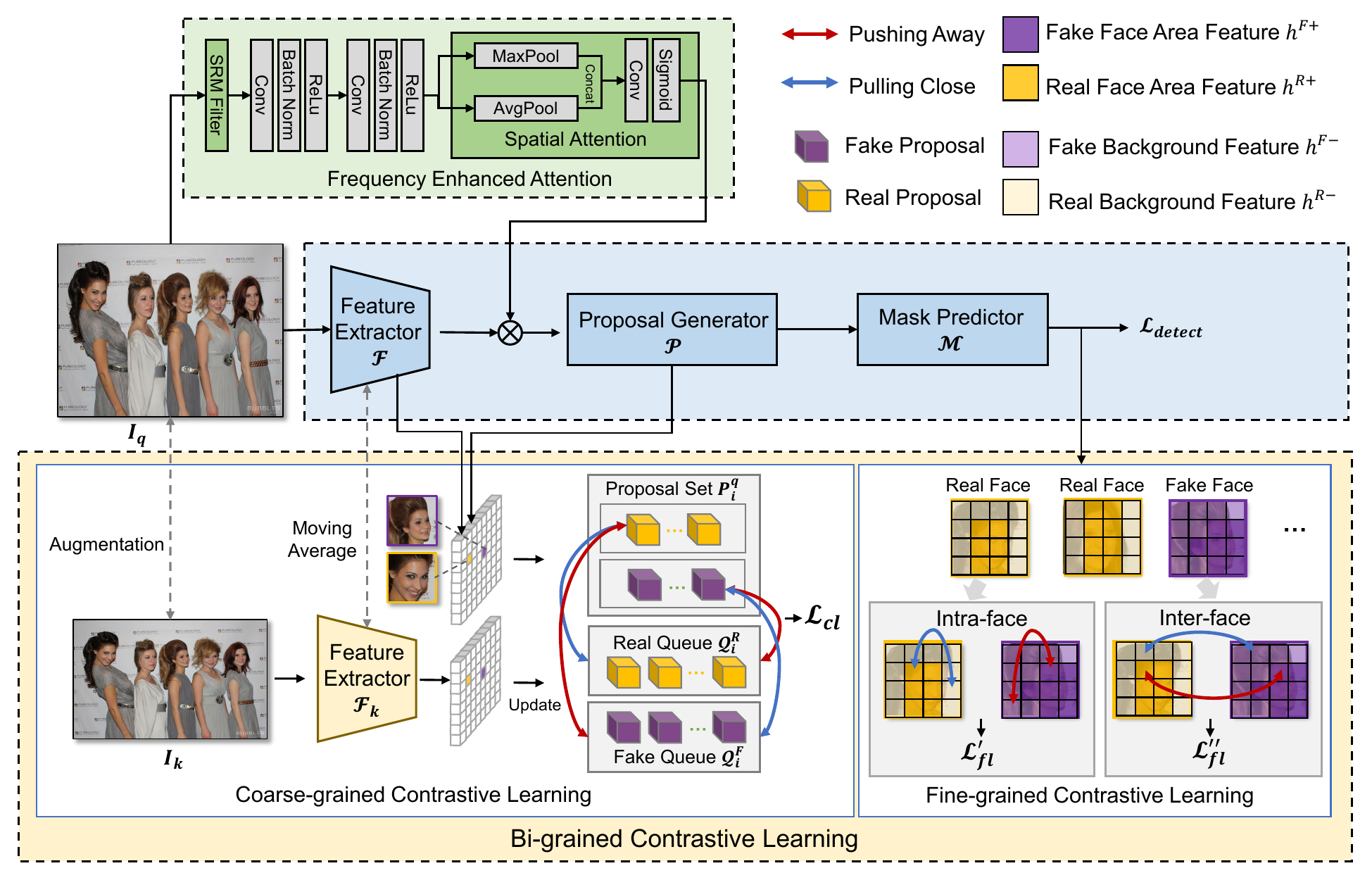} 
        \vspace{-0.2cm}
	\caption{\small Overview of the proposed bi-grained contrastive learning. Our method is designed on the single-stage architecture, containing the feature extractor, the proposal generator, and the mask predictor respectively. Specifically, we perform coarse-grained contrastive learning on a feature extractor with the guidance of a proposal generator to capture the forgery traces among different face proposals, and fine-grained contrastive learning on mask predictor by considering the relationship of pixels in the same face (Intra-face) or different faces (Inter-face).}
	\label{fig:overview}
\end{figure*}

In this paper, we explore a new framework called \underline{Co}ntrastive \underline{M}ulti-FaceForens\underline{ics} (COMICS).
 {This framework can be integrated with various object detectors to enable the detection of multiple face forgeries in a single pass, as illustrated in Figure~\ref{fig:twostage} (right).}
As the key component of COMICS, the aim of the bi-grained contrastive learning is to extract the task-related features from real and fake face pairs, operating at both coarse-grained and fine-grained levels. 
For coarse-grained contrastive learning, we first obtain the feature representations at various layers from the feature extractor corresponding to faces. Considering discrimination in features of different layers, we design contrastive learning among positive and negative proposals within each layer separately, and then the contrastive losses at different layers are combined into a multi-layer contrastive loss, aiming to capture the global discrepancy between real and fake faces. 
After that, fine-grained contrastive learning is devised to obtain local forgery traces, including intra-face and inter-face contrastive learning at the pixel level.  
The intra-face learning aims to capture the pixel-wise discrepancy between the original face and the forged counterparts, while inter-face learning focuses on learning the pixel-wise content inconsistency between real and fake faces.
To boost the discovery of forgery traces among faces, we further design {a} frequency enhanced attention module, which enhances the forged traces with
high-frequency features at multiple feature layers by combining an SRM filter \cite{fridrich2012rich} and spatial attention \cite{woo2018cbam}.
Our framework is trained end-to-end (Fig.~\ref{fig:overview}), facilitating the model to learn sufficiently to {discover} forged traces.

Extensive experiments are conducted on the OpenFaceForensics and FFIW datasets compared to many state-of-the-art methods that are adapted from object detection. The results {corroborate} the superiority of our method in detecting multiple faces forgeries. We also conduct a comprehensive ablation study to explore the effect of different configurations and demonstrate our method can be integrated into various architectures. 
Our contributions are summarized as follows:

\begin{enumerate}
\item[-] We propose Contrastive Multi-FaceForensic (COMICS), an end-to-end contrastive learning framework for multi-face forgery detection in the wild. This learning framework is plug-and-play, and easily adapted to different architectures.

\item[-] We propose a new bi-grained contrastive learning scheme that considers coarse- and fine-grained contrastive learning to capture the multi-scale proposal-wise and pixel-wise forgery trace hierarchically in both inter- and intra-face manner. Additionally, we propose a frequency enhanced attention module aiming at extracting fine-grained forgery traces. 

\item[-] Experimental results on the OpenForensics and FFIW datasets demonstrate the proposed method significantly outperforms other state-of-the-art counterparts. We also study our method regarding many factors, including the effect of each level of contrastive learning, and different parameters settings and component combination, providing more insight into the future research.  

\end{enumerate}


\section{Related Work}
\label{sec:relatedworks}
\noindent{\bf Face Forgery Recognition.}
Most studies follow the two-stage pipeline which extract the face area first and then determines its authenticity. Different types of visual cues are used for the classification, including physiological signals (\eg, eye blinking \cite{li2018ictu}, mouth movements \cite{10023530}, head pose \cite{yang2019exposing,ivanov2020combining}, blood flow and heartbeats \cite{ciftci2020fakecatcher}, domain inconsistency \cite{haliassos2022leveraging,10017352,10018271}, {\it etc.}) and signal abnormalities (\eg, face warping artifacts \cite{li2018exposing}, PRNU difference \cite{koopman2018detection,9795231},  {spatial inconsistency \cite{10025759,10138555}}, multimodal inconsistencies \cite{10233898,10107603,10243082,10233898} and frequency artifacts \cite{frank2020leveraging,durall2019unmasking,song2022adaptive,huang2023dodging}, {\it etc.}). Other methods designed  {specific architectures \cite{afchar2018mesonet,liu2020global,luo2021generalizing,10163252}}, data augmentations \cite{li2018exposing,shiohara2022detecting} and  {self-supervised learning \cite{9903059}} to learn effective features. To further improve detection performance, many datasets are proposed, such as FaceForensics++ \cite{rossler2019faceforensics}, Celeb-DF \cite{li2020celeb}, DFDC \cite{dolhansky2019deepfake}, {\it etc.}. These datasets are constructed under a restricted environment, only containing a single subject in each video. However, the impressive performance on these datasets does not indicate the effectiveness of existing methods under real situations that contain complex background and multiple or even dense faces.


\smallskip
\noindent{\bf Multi-face Forgery Detection.}
To address the above mentioned issues, the OpenForensics dataset \cite{le2021openforensics} was proposed to encourage the research on single-stage methods for multi-face forgery detection. Several advanced object detection models, including two-stage models that detect and then segment (\eg, MaskRCNN \cite{he2017mask}, MSRCNN \cite{huang2019mask}, {\it etc.}), and other single-stage models emphasizing real-time performance (\eg, RetinaMask \cite{fu2019retinamask}, YOLACT \cite{bolya2019yolact}, YOLACT++ \cite{bolya2020yolact}, CenterMask \cite{lee2020centermask}, PolarMask \cite{xie2020polarmask}, MEInst \cite{zhang2020mask}, ConInst \cite{tian2020conditional}, SOLO \cite{wang2020solo}, and SOLO2 \cite{wang2020solov2}), are directly adapted to this task, to detect face area and its authenticity simultaneously. 
Despite these models adapted from object detection domain show the feasibility of detecting multi-face forgery, they are not the optimal solution as they are not designed to capture the subtle forgery traces. In this paper, we propose a new single-stage framework that can capture the traces of DeepFake synthesis using a bi-grained contrastive learning scheme.

\smallskip
\noindent{\bf Contrastive Learning.}
Contrastive learning is one of the most compelling methods in both unsupervised and supervised visual representation learning domains \cite{chen2020simple,he2020momentum,zhang2022enhancing}, which has been widely used in object detection \cite{xie2021detco,li2022class, zhang2022edge}, segmentation \cite{wang2022contrastmask,wang2022uncertainty}, medical image~\cite{liu2022margin}, personalized recommendation\cite{10075502,9712249}, image captioning \cite{dai2017contrastive,li2020context}, and natural language understanding \cite{qu2020coda}. Inspired by contrastive learning,  the work of \cite{chugh2020not} focused on the inconsistency between audio and visual modalities, and used a contrastive loss to model inter-modality similarity for DeepFake video detection.  {MCL \cite{10243082} uses contrastive learning to expose inconsistency clues across frame, video, and audio modalities, but its application is restricted to video content.} The work of \cite{kumar2020detecting} proposed a triplet network to increase the feature distance between real and fake videos in the embedded feature space. CFFN \cite{hsu2020deep} used the siamese network and contrastive loss to learn the discriminative features to classify GAN-generated faces. DCL \cite{sun2022dual} is the most relevant method which considers different instance-level contrastive learning, focusing on the global discrepancy between real and fake faces, as well as the intra-face local forgery traces within individual facial regions.
However, these methods are designed for two-stage frameworks {and do not consider the inconsistency between the real and fake faces in the image}, thus not applicable to single-phase frameworks. 

\section{Contrastive Multi-FaceForensics}
In this paper, we propose Contrastive Multi-FaceForensics (COMICS) dedicated to {capture} the forgery traces among multiple faces in an end-to-end fashion. In contrast to the conventional face forgery recognition methods, our method detects multiple face forgeries simultaneously by considering the coarse-grained and fine-grained levels of contrastive information among faces. 

\smallskip
\noindent{\bf Network Architecture.} The network architecture of our method is derived from the recent object detectors \cite{chen2020blendmask,tian2019fcos,tian2020conditional}. The object detectors are end-to-end architectures that generally contain a feature extractor, a proposal generator, and a mask predictor, respectively. They can locate the objects and give their semantic categories simultaneously. By simply restricting the object categories to two categories of real or fake faces, these architectures can be adapted to multi-face forgery detection, which thus naturally solves the limitations of face forgery recognition methods to a certain extent \cite{le2021openforensics}.  Inspired by that, we adopt such an end-to-end architecture and propose a new bi-grained contrastive learning scheme customized for the feature extractor, proposal generator, and mask predictor. 

\smallskip
\noindent{\bf Bi-grained Contrastive Learning.} Our study is inspired by the general contrastive learning strategy in that the forgery traces may be exposed by inspecting the relationship between faces. Specifically, our method seeks these relationships at the coarse-grained (proposal-wise) and fine-grained (pixel-wise) levels on the end-to-end architecture. The elements in features from the feature extractor correspond to different face proposals across various scales, and the elements in features from the mask predictor correspond to different pixels in faces. Applying bi-grained contrastive learning can boost the network to capture the forgery traces at multi-scale coarse- and fine-grained levels, achieving the goals of simultaneously detecting faces and their authenticity. Fig.~\ref{fig:overview} provides the overview of the proposed framework,  {while the corresponding notations and definitions are detailed in the Appendix.} Next, Sec.~\ref{sec:coarse} and Sec.~\ref{sec:fine} introduce the details of coarse-grained and fine-grained contrastive learning, respectively.

\begin{figure*}[t]
	\centering
	\includegraphics[width=0.7\linewidth]{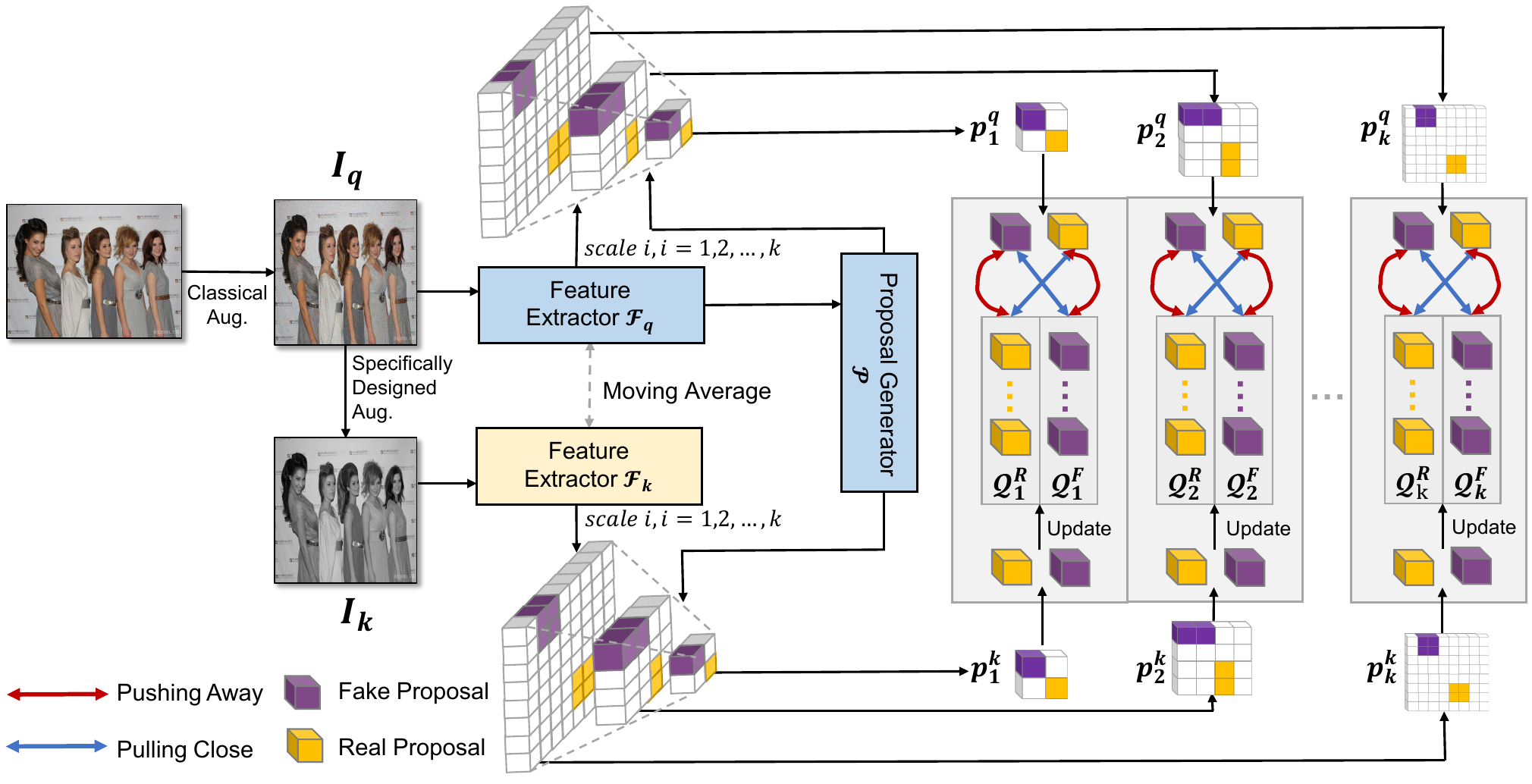} 
 	\vspace{-0.4cm}
	\caption{\small  {Overview of the coarse-grained contrastive learning. The input image is first augmented into two views ($I_q,I_k$) and then contrastive learning is performed at the proposal levels on different scales of the feature extractor. Note that the feature elements (\eg, the yellow or purple blocks) correspond to real or fake faces given the proposal generator $\mathcal{P}$. }}
	\label{fig:coarse} 
\end{figure*}

\subsection{Coarse-grained Contrastive Learning}
\label{sec:coarse}
The coarse-grained contrastive learning focuses on capturing the discriminative forgery features among positive and negative proposals based on the feature extractors, \ie, pulling close the features of proposals in the same category and pushing away the ones of different categories. For an input image, we generate two samples of different views and design a pair of feature extractors for contrastive learning, where the two feature extractors share the same network architecture but with different parameters. Then these two samples are respectively fed into the feature extractors to obtain proposals at various layers. Since proposals from different layers are not directly comparable, we conduct contrastive learning at each layer separately to capture multi-layer proposal-wise forgery traces by matching the proposals with faces. Fig.~\ref{fig:coarse} shows the detail of coarse-grained contrastive learning. In what follows, we introduce the data view configuration and contrastive learning configuration, respectively.

\subsubsection{Data View Configuration}
In general classification tasks, classical data augmentation methods (\eg, random crop, random flip, random rotation, {\it etc.}) are used to create different data views for contrastive learning. Since the main content is inside the image, using these augmentations can retain the main content by large and enable the network to learn the content-relevant features. 
However, those general augmentations are unsuitable for our task as they possibly disrupt the location of faces, because the position of faces between two views should be matched to establish correspondence among faces for contrastive learning (see Fig.~\ref{fig:coarse}). Face forgery detection is less influenced by image color and structure than other vision tasks, so we employ several data augmentations specifically designed to mitigate the impact of task-irrelevant visual attributes.  {These include color jittering, grayscale conversion, random blocking, bilinear interpolation, Gaussian noise, and salt-and-pepper noise. The detailed settings for these augmentations are provided in the Experimental Settings section.}

These augmentation methods well simulate the diversity existing in real world data, while they have a negligible influence on the position of faces. The data augmentation provide richer samples for our method to learn the true visual forgery traces.
Given an image $I\in \mathbbm{R}^{H\times W\times3}$, we first utilize the classical data augmentations, {\it i.e.}, random crop, random flip, and random rotation, to generate $I_{q}$. Then we employ the above four kinds of specifically designed data augmentations to generate the other view $I_{k}$.

\subsubsection{Contrastive Learning Configuration} In the contrastive learning context, MOCO \cite{he2020momentum}, a well-known learning paradigm, utilizes image pairs to build dynamic dictionaries for contrastive learning to extract discriminative visual representations. To extract the face forgery traces with contrastive learning, we build a similary contrastive learning pipeline as MOCO. Denote the feature extractor as $\mathcal{F}_{q}$ with parameters $\theta_{q}$, we build the same feature extractor $\mathcal{F}_{k}$ with parameters ${\theta}_{k}$.
The feature extractor $\mathcal{F}_{k}$ is updated as
\begin{equation}
    \theta_{k} =\beta \cdot \theta_{k} + (1-\beta) \cdot {\theta}_{q},
\end{equation}
where $\beta$ is the hyper-parameter to control the update ratio. 

By feeding the image $I_q$ into feature extractor $\mathcal{F}_{q}$, we obtain a set of features as $\{f_{1}^{q},...,f_{i}^{q},..., f_{n}^{q} \} = \mathcal{F}_{q}(I_q)$. Specifically, we utilize the features extracted from the $3,4,5,6$ and $7$-th layers. We denote the proposal generator as $\mathcal{P}$. These features are then sent into $\mathcal{P}$ to create proposals on each feature as $\{P_{1}^{q},..., P_{n}^{q} \} = \mathcal{P}(\mathcal{F}_{q}(I_q))$, where $P_{i}^{q}$ denotes a set of proposals generated on the $i$-th feature. Similarly, for image $I_k$, we can obtain a set of features from the feature extractor $\mathcal{F}_{k}$ as $\{f^{k}_{1},...,f^{k}_{i},..., f^{k}_{n} \} = \mathcal{F}_{k}(I_k)$. These features are then sent into the same proposal generator $\mathcal{P}$ to create corresponding proposals as $\{P^{k}_{1},...,P^{k}_{1},..., P^{k}_{n} \} = \mathcal{P}(\mathcal{F}_{k}(I_k))$.


\smallskip
\noindent{\bf Inter-faces.} For a specific layer of $i$, we can obtain two proposal sets $P_{i}^{q}$ and $P^{k}_{i}$ corresponding to the input images $I_q$ and $I_k$. The proposal is viewed as real (fake) if IoU between this proposal and an arbitrary real (fake) face is greater than $0.6$. Then we find the corresponding feature representation of each proposal from the feature extractors. For contrastive learning, we design two queues, a real face queue $\mathcal{Q}^{R}_i$ and a fake face queue $\mathcal{Q}^{F}_i$ to store the feature representations of the corresponding proposals. These queues are updated dynamically in training. Denote $p_{ij}^{q}$ and $p^{k}_{ij}$ as the feature representation of the $j$-th proposal at the $i$-th layer in $P^{q}_{i}$ and $P^{k}_{i}$ respectively. $\hat{p}_{i}^{R}$ and $\hat{p}_{i}^{F}$ are two prototypes used to update the real face queue $\mathcal{Q}^{R}_i$ and the fake face queue $\mathcal{Q}^{F}_i$. If $p^{k}_{ij}$ corresponds to a fake proposal, the prototype $\hat{p}_{i}^{F}$ is updated as
\begin{equation}
    \hat{p}^{F}_{i} = \alpha \cdot  \hat{p}^{F}_{i}+(1-\alpha) \cdot p^{k}_{ij},
\end{equation}
where $\alpha$ is a factor controlling the update process, similar to \cite{sun2022dual}.
To update $\mathcal{Q}^{F}_i$, we calculate the similarity between $\hat{p}^{F}_{i}$ and $p^{k}_{ij}$ in a batch, and push the features of top-5 dissimilar proposals into $\mathcal{Q}^{F}_i$. The update scheme for $\hat{p}_{i}^{R}$ and $\mathcal{Q}^{R}_i$ are the same if $p^{k}_{ij}$ corresponds to a real proposal.
The similarity between the two proposals is measured by cosine similarity
\begin{equation}
    \delta(p^{q}_{ij},p^{k}_{ij})= \frac{p^{q}_{ij}}{\|p^{q}_{ij}\|}\cdot\frac{p^{k}_{ij}}{\|p^{k}_{ij}\|}
\end{equation}

Note that on a certain layer, the number of proposals may be small. We use FlatNCE \cite{chen2021simpler} to eliminate the floating-point overflow caused by the insufficient number of samples in the training batch. The objective at the $i$-th layer is defined as

\begin{equation}
\mathcal{L}_{cl}^{i} = \frac{\sum_{p^{k}_{ih}\in \mathcal{Q}^{-}} 
e^{\delta(p^{q}_{ij},p^{k}_{ih})/\tau-\delta(p^{q}_{ij},p^{k}_{ij})/\tau}}
{\texttt{Detach}(\sum_{p^{k}_{ih}\in \mathcal{Q}^{-}} 
e^{\delta(p^{q}_{ij},p^{k}_{ih})/\tau-\delta(p^{q}_{ij},p^{k}_{ij})/\tau})},
\label{eq:coai}
\end{equation}
where $\texttt{Detach}(\cdot)$ is an operation that stops the floating point relative error to be back-propagated by  gradients, and $\tau$ is the temperature parameter. $\mathcal{Q}^{-}$ denotes the negative queue corresponding to $p_{ij}$, where $\mathcal{Q}^{-}=\mathcal{Q}^{F}_{i}$ when $p_{ij}$ corresponds to a real face, otherwise $\mathcal{Q}^{-}=\mathcal{Q}^{R}_{i}$. 

\smallskip
\noindent{\bf Multi-layer Ensemble.} Since different layers represent different levels of visual information, we extend Equ.~(\ref{eq:coai}) to multi-layer contrastive loss as

\begin{equation}
    \mathcal{L}_{cl}= \sum_{i} \omega_{i} \cdot \mathcal{L}_{cl}^{i},
    \label{eqn:multiscale}
\end{equation}
where $\omega_{i}$ is the weight factor for different layers. 
The coarse-grained contrastive learning uses the specific designed data augmentations to learn the informative and discriminative visual attributes for face forgery detection at different layers. 

\subsection{Fine-grained Contrastive Learning}
\label{sec:fine}
\begin{figure}[t]
	\centering
	\includegraphics[width=\linewidth]{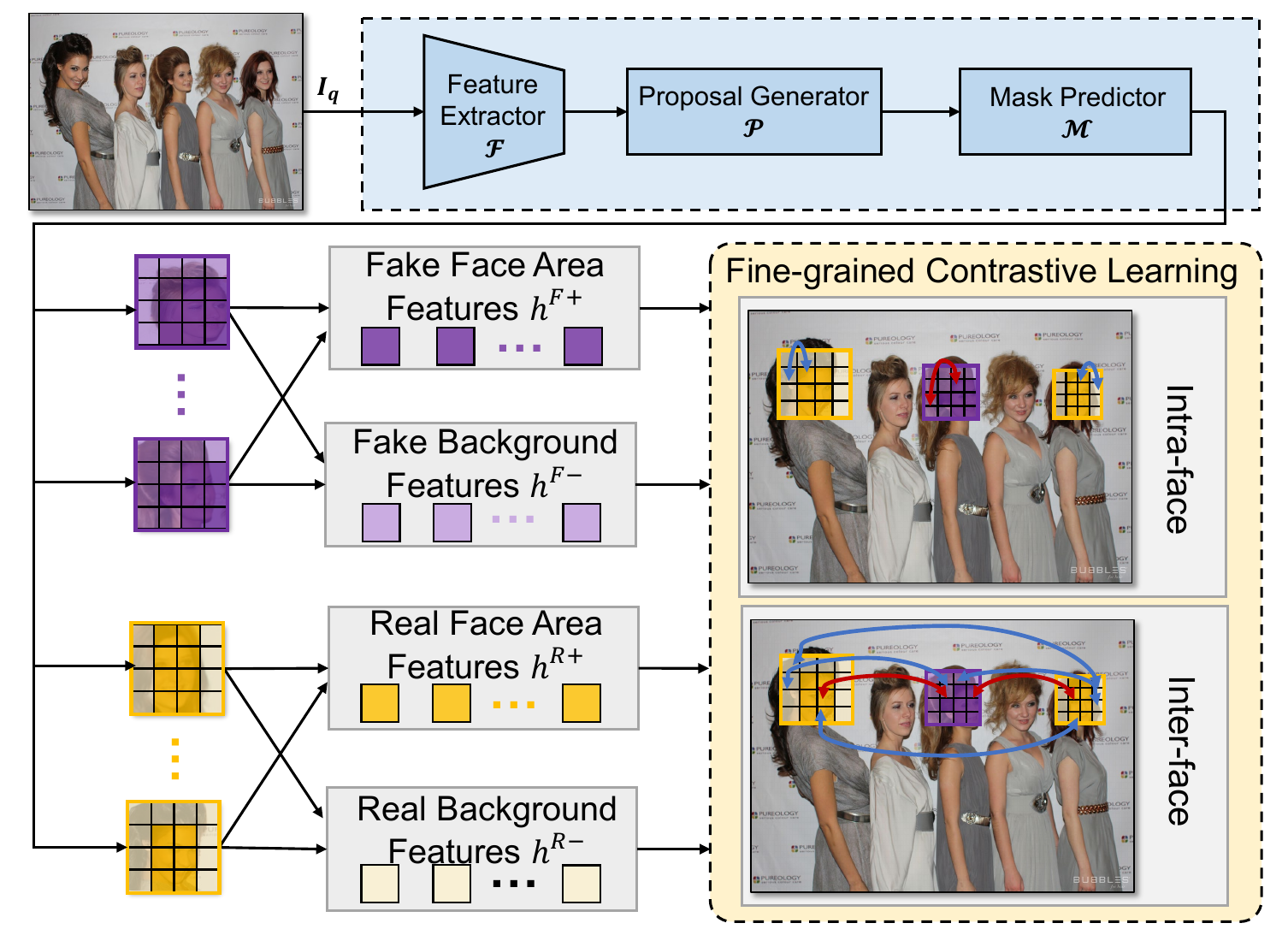} 
 	\vspace{-0.7cm}
	\caption{\small Overview of the fine-grained contrastive learning. In contrast to coarse-grained contrastive learning, it aims to learn the relationship among pixels instead of proposals. Given the feature maps of the predicted masks, we consider both intra-face and inter-face relations. Specifically, the intra-face relation aims to capture the inconsistency between the forged and the surrounding original area in the same face, while inter-face relation explores the pixel-wise discrepancy between different faces. }
	\label{fig:fine}
  \vspace{-0.5cm}
\end{figure}

Besides the coarse-grained level, we also consider fine-grained contrastive learning, which focuses on the pixel-wise inter-face and inter-face relationship. For the intra-face relation, we aim to learn the inconsistency between the forged and the original surrounding area in the same face. For the inter-face relation, we explore the pixel-wise discrepancy between faces. The overview of fine-grained contrastive learning is shown in Fig.~\ref{fig:fine}.

Denote the mask predictor as $\mathcal{M}$. Given the above proposals, we can obtain the corresponding feature maps of the predicted mask as $\{h_1,...,h_m \}=\mathcal{M}(\mathcal{P}(\mathcal{F}_{q}(I_{q})))$. Here we mix the proposals of different layers into one set, where $h_j$ is the feature map of the $j$-th predicted mask. Then we resize the feature map and the corresponding predicted mask into a fixed size.
Denote the feature maps of predicted masks corresponding to real and fake faces as $\mathcal{H}^R$ and $\mathcal{H}^F$, respectively. Note that a mask always covers the face area and the background area. For $h_j \in \mathcal{H}^R$, both the face part and the background part are the original visual content. However, for $h_j \in \mathcal{H}^F$, the face part is manipulated while the background remains original. To learn the intra- and inter-face forgery traces at pixel level, we split each predicted mask into two groups. Specifically, if $h_j \in \mathcal{H}^R$, we use $h^{R+}_j$ and $h^{R-}_j$ to denote the feature maps of the face area and the background, otherwise they are denoted as $h^{F+}_j$ and $h^{F-}_j$. 

\begin{figure}[t]
	\centering
	\includegraphics[width=0.6\linewidth]{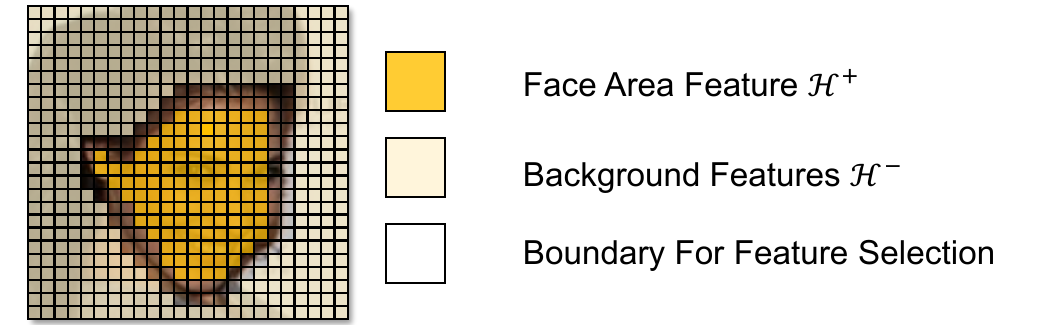} 
	\caption{\small Illustration of mask split. In our method, we discard the pixels that have a distance within two pixels away from the boundary.\textit{ Note that we use transparent blocks to denote the face boundary.}}
	\label{fig:face}
\end{figure}

Standard face forgery methods usually apply postprocessing on the boundary to remove artifacts, which 
brings about difficult in recognizing the forgery features compared to other pixels.
Due to the overediting on the face boundaries, we do not take the pixels close to the forgery boundary into consideration. Instead, we discard the pixels that have a small distance within two pixels away from the boundary to guarantee the confidence coefficient of the real and fake examples. The split of a mask is illustrated in Fig.~\ref{fig:face}. We introduce intra-face and inter-face contrastive learning as follows. 

\smallskip
\noindent{\bf Intra-face.}
For a fake face, the features of the forged area should be different from the ones of the background, thus we maximize the distance between these two areas. However, for a real face, these features should be similar. Therefore, We minimize the distance between these two areas. The intra-face learning can be formulated as
\begin{equation}
    \small
    \mathcal{L}'_{fl} =
     - \log \frac{\sum\limits_{h_j \in \mathcal{H}^R} e^{\delta\left (h_{j}^{R+} ,h_{j}^{R-} \right ) /\tau } }
{\sum\limits_{h_j \in \mathcal{H}^R} e^{\delta\left ( (h_{j}^{R+},h_{j}^{R-} \right )/\tau } + \sum\limits_{h_j \in \mathcal{H}^F}  e^{\delta\left ( h_{j}^{F+},h_{j}^{F-} \right ) /\tau } },
\end{equation} 
where $\delta(\cdot)$ is the normalized cosine similarity between two regions
\begin{equation}
    \delta(h ,h') = \sum_{o \in h} \sum_{o' \in h'} \frac{o}{\|o\|}\cdot\frac{o'}{\|o'\|},
\end{equation}
where $o$ denotes the feature representation of one pixel. $\tau$ is the temperature parameter.


\smallskip
\noindent{\bf Inter-face.}
Different from intra-face contrastive learning that considers the content discrepancy inside a face, inter-face learning focuses on exploring the relationship between different faces. Since the backgrounds of both real and fake faces are the original visual content, we pull the features of the backgrounds between real and fake faces to be close. On the contrary, the face area of real and fake faces should be different, thus we push away their corresponding features. The inter-face contrastive learning can be defined as
\begin{equation}
\small
    \mathcal{L}''_{fl} =  - \log \frac{\sum\limits_{h_{k}\in \mathcal{H}^{-}} e^{\delta\left ( h_{j}^{R-},h_{k}^{F-} \right ) /\tau } }
{\sum\limits_{h_{k}\in \mathcal{H}^{-}} e^{\delta\left ( h_{j}^{R-},h_{k}^{F-} \right )/\tau } + \sum\limits_{h_{k}\in \mathcal{H}^{-}} e^{\delta\left (  h_{j}^{R+},h_{k}^{F+}  \right ) /\tau } },
\end{equation}
where $\mathcal{H}^{-}$ denotes the negative set of $h_{j}$, \ie, $\mathcal{H}^{-}$ corresponds to $\mathcal{H}^{R}$ if $h_{j}$ is fake, otherwise $\mathcal{H}^{F}$.
After performing fine-grained contrastive learning from both inter-face and intra-face aspects, the discriminative information between real and fake parts in the same images can be fully captured.

\section{Frequency Enhanced Attention and Overall Objectives}


\noindent{\bf Frequency Enhanced Attention.}
Inspired by the previous works \cite{zhou2023discriminative,chen2021image,fei2022learning} that frequency feature can effectively expose forgery traces, we design a simple frequency enhanced attention to enhance the importance of features indicated by high-frequency elements. 
We employ the spatial rich model (SRM) filter \cite{fridrich2012rich} on the input images and features of intermediate layers to focus on the high-frequency components.

As shown in the top left part of Fig.~\ref{fig:overview}, we construct a new branch that contains two convolutional blocks, where each convolutional layer is followed by a Batch Normalization (BN) \cite{ioffe2015batch}, a ReLu activation. Then a spatial attention layer \cite{woo2018cbam} is designed to receive the output of ReLu layer of the second convolutional block. The output of this branch will be multiplied by the output of the feature extractor. 
By employing frequency enhanced attention, the difference between the real and fake features is amplified, which contributes to enhancing the discriminative power of the features.

\smallskip
\noindent{\bf Overall Objective Function.} 
We first adapt the vanilla object detection objectives to our task by simply modifying the semantic object categories to the category of real and fake faces. We denote the objective loss as $\mathcal{L}_{detect}$.

By further considering coarse-grained and fine-grained contrastive learning, the objectives of bi-grained contrastive learning can be written as 
\begin{equation}
    \mathcal{L}_{Bi}=\lambda_{1} \mathcal{L}_{cl}+ \lambda_{2} \mathcal{L}'_{fl} + \lambda_{3} \mathcal{L}''_{fl},
\label{equ:loss}
\end{equation}
where $\lambda_{1}$, $\lambda_{2}$, and $\lambda_{3}$ are the hyper-parameters for the components of bi-grained contrastive learning. Together with $\mathcal{L}_{detect}$, we can train the multi-face forgery detection model end-to-end (Fig.~\ref{fig:overview}).


\noindent{\bf Discussion.} COMICS is plug-and-play and can be adapted to various backbone networks. Different from previous works DCL \cite{sun2022dual} and CFFN \cite{hsu2020deep} that model face features at the single feature layer, our method explicitly models and combines the relationships between faces in the same image and compares proposals at different layers, encouraging a more elegant forgery feature learning, leading to improved detection rates of fake faces and clearer face boundaries.
Towards a wider impact, our work may inspires the development of contrastive learning on vision tasks that require the discovery of more fine-grained features. For example, the bi-grained and multi-layer contrastive learning can be applied to object detection on fine-grained visual object categories. Our data augmentation strategy may also provide insights for aligning features of different views in multi-view visual learning, benefiting tasks such as aligning infrared and depth with RGB images.



\section{Experiments}
\label{sec:exps}
\subsection{Experimental Settings}
\begin{figure}[t]
	\centering
	\includegraphics[width=\linewidth]{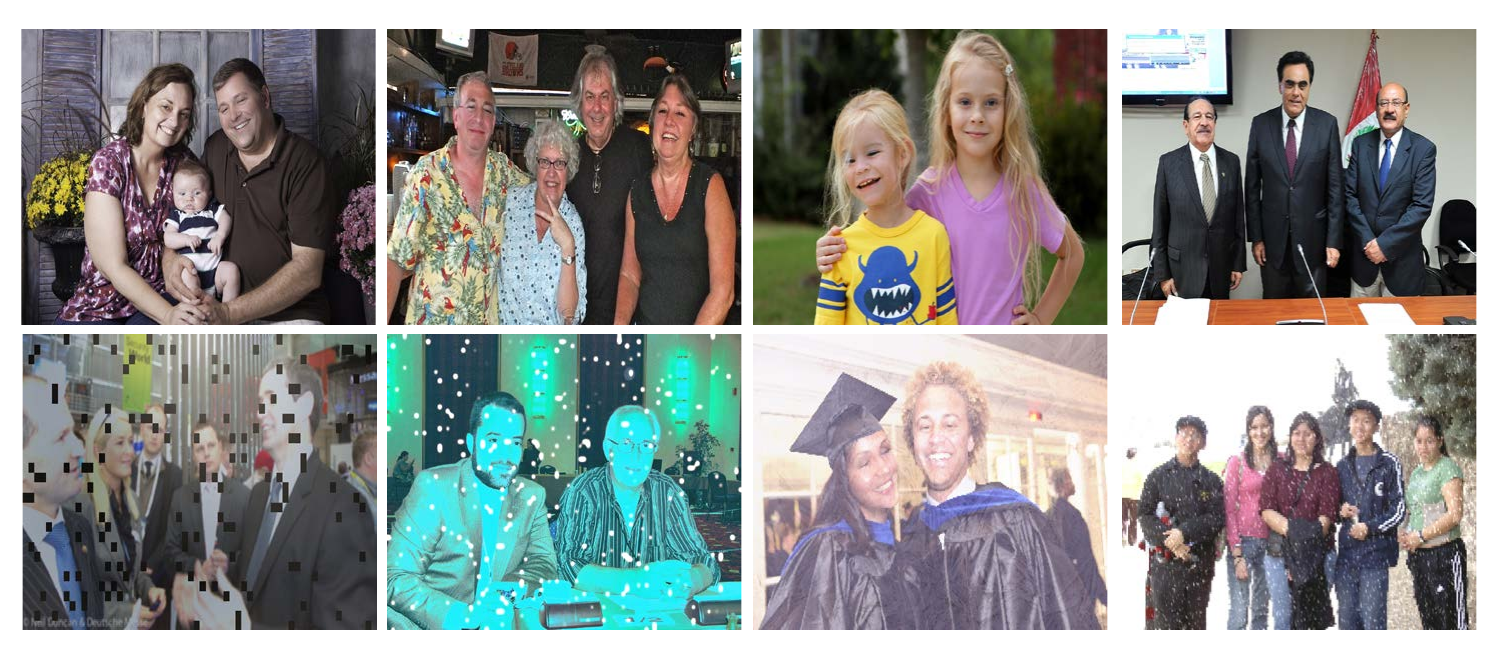} 
 	\vspace{-0.7cm}
	\caption{\small Image examples of OpenForensics dataset. The first row is selected from test-development set and the second row is selected from test-challenge set.}
	\label{fig:dataset}
\end{figure}

\noindent{\bf Dataset.} Most of the experiments are performed on OpenForensics \cite{le2021openforensics}, a recently released large-scale dataset of face forgeries. It has rich annotations including class (real or fake), face forgery location and mask, which is suitable for multi-face forgery detection and segmentation. The training set contains $44,122$ images with $85,392$ real faces and $65,972$ fake faces. The validation set contains $7,308$ images including $4,786$ real faces and $10,566$ fake faces. This dataset has two testing sets according to the challenge level of detection, which are the test-development set and the test-challenge set. The images in test-development have the same distribution as the training set, while the images in the test-challenge set are processed by many operations such as block-wise distortion, color manipulation, and random noise, see examples in Fig.~\ref{fig:dataset}. The test-development set has $18,895$ images with $21,071$ real faces and $28,670$ fake faces, while the test-challenge set has $45,000$ images with $49,218$ real faces and $68,452$ fake faces. 
Note that the number of datasets for multi-face forgery detection is significantly less than the number of datasets for conventional face forgery recognition. We also evaluate the proposed method on the FFIW dataset \cite{Zhou_2021_CVPR}, which is a DeepFake video dataset that contains both real and fake faces within the same video, but only includes ground truth for fake faces while the locations of real faces remain unknown. 
 {The FFIW dataset contains $16,000$ videos for training, $500$ videos for evaluation, and $3476$ videos for testing.} For comparison, we extract the frames for every two seconds of the video and get a dataset of $126,360$ images for training, $31,240$ images for testing, and $3,560$ images for evaluation.

\smallskip
\noindent{\bf Compared Methods.} Following the settings in \cite{le2021openforensics}, we compare the proposed
methods with several state-of-the-art object detection methods including MaskRCNN \cite{he2017mask}, MSRCNN \cite{huang2019mask}, RetinaMask \cite{fu2019retinamask}, YOLACT \cite{bolya2019yolact}, YOLACT++ \cite{bolya2020yolact}, CenterMask \cite{lee2020centermask}, BlendMask \cite{chen2020blendmask}, PolarMask \cite{xie2020polarmask}, MEInst \cite{zhang2020mask}, CondInst \cite{tian2020conditional}, SOLO \cite{wang2020solo}, and SOLO2 \cite{wang2020solov2}. {Besides, we adopt the DCL \cite{sun2022dual} framework on BlendMask, denoted as \textit{BlendMask+DCL}, which conducts contrastive learning between real and fake faces, as well as the intra-face local forgery traces within individual facial regions.}



\smallskip
\noindent{\bf Evaluation Metrics.} Following the same settings in \cite{le2021openforensics}, we use the COCO-style Average Precision (AP) and Localization Recall Precision (LRP) for evaluation {on face forgery detection and segmentation. The prediction accuracy is determined by the IoU threshold and ground truth for face forgery detection and segmentation. }
Specifically, we evaluate the results using mean AP, AP$_{S}$, AP$_{M}$, and AP$_{L}$, where $S$, $M$, and $L$ represent small, medium, and large objects, respectively. For LRP, we present results for mean optimal LRP (oLRP), localization (oLRP$_{Loc}$), false positive rate (oLRP$_{FP}$), and false negative rate (oLRP$_{FN}$). The higher AP and lower oLRP indicate better performance. In this section, we discuss the performance gain in absolute values.

\begin{table*}[t]
	\centering
	\small
	\caption{\small Comparisons of our method with others on the OpenForensics test-development set. Higher AP is better and lower oLRP error is better.}
        \vspace{-0.3cm}
        \resizebox{\linewidth}{!}{
	\begin{tabular}
  {l|cccc|p{1cm}<{\centering}p{1cm}<{\centering}p{1cm}<{\centering}p{1.1cm}<{\centering}|cccc|p{1cm}<{\centering}p{1cm}<{\centering}p{1cm}<{\centering}p{1.1cm}<{\centering}}
  \hline
	 \multirow{2}{*}{Method}    & \multicolumn{8}{c|}{Multi-Face Forgery Detection} & \multicolumn{8}{c}{Multi-Face Forgery Segmentation}  \\
            \cline{2-17}
            & AP & AP$_{S}$$ $ & AP$_{M} $ & AP$_{L} $ &  oLRP &  oLRP$_{Loc}$ &oLRP$_{FP}$ & oLRP$_{FN} $ & AP  &AP$_{S} $ & AP$_{M}$ & AP$_{L}$ & oLRP & oLRP$_{Loc} $ & oLRP$_{FP} $ & oLRP$_{FN} $ \\
  		\hline 
        MaskRCNN \cite{he2017mask}    & 79.2 & 29.9 & 80.2 & 79.5 & 24.3 & 9.5 & 2.7 & 4.0 & 83.6 & 16.1 & 82.1 & 85.8 & 21.2 & 7.6 & 3.0 & 4.2 \\
        MSRCNN   \cite{huang2019mask} & 79.0 & 29.5 & 80.1 & 79.5 & 24.3 & 9.6 & 2.7 & 3.8 & 85.1 & 16.8 & 84.2 & 86.8 & 21.1 & 7.7 & 2.6 & 4.4 \\
        \hline 
        RetinaMask \cite{fu2019retinamask} & 80.0 & 30.9 & 80.2 & 80.7 & 24.2 & 9.0 & 3.0 & 4.6 & 82.8 & 16.4 & 80.6 & 85.1 & 22.6 & 8.1 & 2.9 & 4.9 \\
        YOLACT      \cite{bolya2019yolact} & 68.1 & 12.5 & 67.1 & 69.3 & 37.2 & 13.4 & 6.3 & 8.7 & 72.5 & 3.1 & 67.0 & 75.7 & 34.0 & 11.4 & 6.4 & 8.7 \\
        YOLACT++    \cite{bolya2020yolact} & 85.0 & 27.4 & 85.4 & 85.7 & 20.7 & 6.6 & 2.5 & 6.6 & 85.0 & 15.3 & 83.3 & 87.0 & 21.3 & 6.9 & 2.5 & 6.6 \\
        CenterMask \cite{lee2020centermask}& 85.5 & 32.0 & 85.2 & 86.2 & 21.1 & 6.8 & 3.3 & 5.9 & 87.2 & 16.5 & 85.0 & 89.4 & 21.4 & 6.1 & 3.2 & 7.8 \\
	 
        PolarMask  \cite{xie2020polarmask}& 85.0 & 27.4 & 85.4 & 85.7 & 20.7 & 6.6 & 2.5 & 6.6 & 85.0 & 15.3 & 83.3 & 87.0 & 21.3 & 6.9 & 2.5 & 6.6 \\
        MEInst     \cite{zhang2020mask}& 82.8 & 26.0 & 82.7 & 83.4 & 23.8 & 7.6 & 4.1 & 6.8 & 82.2 & 13.9 & 81.5 & 83.3 & 25.0 & 8.1 & 4.0 & 7.2 \\
        ConInst   \cite{tian2020conditional} & 84.0 & 29.4 & 83.6 & 84.8 & 20.8 & 7.4 & \bf2.3 & 5.2 & 87.7 & 18.1 & 85.1 & 89.8 & 18.3 & 5.9 & 2.4 & 5.3 \\
	  SOLO      \cite{wang2020solo}& - & - & - & - & - & - & - & - & 86.6 & 15.4 & 85.6 & 88.4 & 20.0 & 6.6 & \bf2.1 & 6.0 \\
        SOLO2     \cite{wang2020solov2} & - & - & - & - & - & - & - & - & 85.1 & 13.7 & 83.7 & 87.1 & 21.5 & 7.1 & 3.1 & 5.8  \\
         BlendMask \cite{chen2020blendmask} & 87.0 & \bf 32.7 & \bf86.3 & 88.0 & 19.5 & \bf6.2 & 2.4 & 6.2 & 89.2 & \bf 19.8 & 87.3 & 91.0 & 18.3 & 5.4 & 2.5 & 6.3 \\
     {BlendMask+DCL\cite{sun2022dual}}  & 87.9 & 32.3 & 85.6 & 89.7 & 18.0 & 6.3 & 3.2 & 3.5 & 92.0 & 16.8 & 88.8 & 94.3 & 15.4 & 4.7 & 3.3 & 3.8 \\

   \hline 
       BlendMask+COMICS &  \bf 88.2 & 32.3 & 85.7 & \bf 89.9& \bf 17.4 & \bf 6.2 & 2.5 & \bf3.5 & \bf92.3 & 16.3 & \bf89.1 & \bf94.6 & \bf14.7 & \bf4.7 & 2.6 & \bf3.6 \\

		\hline
	\end{tabular}}
	\label{tab:resultsdev}
 \vspace{-0.3cm}
\end{table*}

\begin{table*}[!th]
	\centering
	\small
	\caption{\small Comparisons of our method with others on the OpenForensics test-challenge set. Higher AP is better and lower oLRP error is better.}
        \vspace{-0.3cm}
        \resizebox{\linewidth}{!}{
	\begin{tabular}
   {l|cccc|p{1cm}<{\centering}p{1cm}<{\centering}p{1cm}<{\centering}p{1.1cm}<{\centering}|cccc|p{1cm}<{\centering}p{1cm}<{\centering}p{1cm}<{\centering}p{1.1cm}<{\centering}}
  \hline
	 \multirow{2}{*}{Method}    & \multicolumn{8}{c|}{Multi-Face Forgery Detection} & \multicolumn{8}{c}{Multi-Face Forgery Segmentation}  \\
            \cline{2-17} 
  		 & AP & AP$_{S}$$ $ & AP$_{M} $ & AP$_{L} $ &  oLRP &  oLRP$_{Loc}$ &oLRP$_{FP}$ & oLRP$_{FN} $ & AP  &AP$_{S} $ & AP$_{M}$ & AP$_{L}$ & oLRP & oLRP$_{Loc} $ & oLRP$_{FP} $ & oLRP$_{FN} $ \\
  		\hline   
    MaskRCNN \cite{he2017mask} & 42.1 & 11.8 & 46.2 & 40.5 & 65.4 & 13.6 & 29.3 & 40.0 & 43.7 & 4.7 & 44.3 & 44.0 & 64.4 & 11.8 & 29.4 & 41.2 \\
    MSRCNN \cite{huang2019mask} & 42.2 & 11.8 & 45.9 & 40.8 & 65.3 & 13.7 & 29.6 & 39.9 & 43.3 & 5.2 & 44.6 & 43.5 & 64.1 & 11.8 & 30.4 & 39.6 \\   
    \hline 
    RetinaMask \cite{fu2019retinamask} & 48.5 & 12.8 & 51.0 & 48.1 & 63.3 & 12.6 & 33.2 & 34.6 & 48.0 & 4.7 & 46.5 & 49.7 & 63.3 & 11.8 & 30.9 & 38.0 \\
    YOLACT \cite{bolya2019yolact}  & 49.4 & 5.6 & 49.6 & 50.3 & 60.1 & 15.3 & 23.2 & 29.9 & 51.8 & 1.4 & 47.2 & 54.6 & 58.4 & 13.5 & 23.4 & 30.1 \\
    YOLACT++ \cite{bolya2020yolact}  & 51.7 & 12.3 & 53.2 & 51.5 & 60.4 & 10.7 & 24.6 & 39.5 & 52.7 & 5.3 & 54.1 & 37.6 & 60.2 & 10.4 & 24.7 & 39.5 \\
    CenterMask \cite{lee2020centermask} & 0.03 & 0.4 & 0.0 & 0.0 & 99.5 & 29.7 & 97.7 & 97.9 & 0.02 & 0.0 & 0.0 & 0.0 & 99.6 & 28.3 & 97.9 & 98.4 \\
 
    PolarMask \cite{xie2020polarmask}& 51.7 & 12.3 & 53.2 & 51.5 & 60.4 & 10.7 & 24.6 & 39.5 & 52.7 & 5.3 & 54.1 & 37.6 & 60.2 & 10.4 & 24.7 & 39.5 \\
    MEInst  \cite{zhang2020mask}& 46.1 & 8.6 & 49.9 & 44.9 & 65.9 & 12.4 & 34.6 & 39.7 & 46.0 & 3.8 & 49.0 & 45.2 & 66.2 & 12.6 & 34.8 & 39.8 \\
    ConInst \cite{tian2020conditional}& 52.7 & 12.6 & 55.3 & 51.8 & 60.7 & 11.5 & 28.3 & 35.3 & 54.1 & 6.5 & 55.2 & 53.8 & 59.6 & 10.0 & 26.7 & 37.3 \\  
    SOLO \cite{wang2020solo} & - & - & - & - & - & - & - & - & 55.9 & 3.9 & 53.3 & 57.3 & 57.6 & 11.3 & 24.6 & 33.0  \\
    SOLO2 \cite{wang2020solov2} & - & - & - & - & - & - & - & - & 53.2 & 3.6 & 52.1 & 54.0 & 59.6 & 11.0 & 24.5 & 37.2  \\
   BlendMask \cite{chen2020blendmask} & 53.9 & 13.5 & 56.6 & 53.5 & 60.2 & 10.6 & 26.5 & 37.4 & 54.0 & \bf 7.1 & 54.5 & 54.5 & 59.9 & 9.8 & 26.4 & 38.4 \\
    {BlendMask+DCL} \cite{sun2022dual}& 66.4 & 16.6 & 65.0 & 67.6 & 46.6 & 9.4 & 16.7 & 24.1 & 67.2 & 6.3 & 61.9 & 70.1 & 45.8 & 8.4 & 17.1 & 24.4 \\ 
  
    \hline 
     BlendMask+COMICS & \bf 74.6 & \bf 19.2 & \bf 72.1 & \bf 76.4& \bf 36.5 & \bf 9.0 & \bf 10.0 & \bf 16.9 &\bf 76.6 &\bf  7.1 & \bf 71.6 & \bf 79.7& \bf 36.3 & \bf 7.9  &\bf  9.6 &\bf  17.9\\

    \hline
	\end{tabular}}
	\label{tab:resultscha}
\end{table*}

\smallskip
\noindent{\bf  {Data Augmentation Details.}}  {For color jittering, we vary the enhancement factors within the ranges from $0$ to $3.1$ for saturation and sharpness, and from $1.0$ to $2.1$ for brightness and contrast. Grayscale conversion transforms RGB images into gray images. The random blocking partitions the image into $10\times10$ chunks and randomly blocks $2\sim6\%$ of the chunks to emphasize the forgery clues. We set the mean to $0$, and select the variance from $[0.01, 0.02, 0.03, 0.04, 0.05]$ for Gaussian noise. The percentage of the added noise for salt-and-pepper noise is selected from $[0.05, 0.1, 0.15]$. For bilinear interpolation, we downscale the image to $1/4$ of its original size and interpolate it back to its original dimensions.}

\smallskip
\noindent{\bf Implement Details.} All experiments were performed on a computing server equipped with one NVIDIA GeForce RTX 3090 GPU. The base architecture is BlendMask \cite{chen2020blendmask}, which uses FPN-ResNet50 \cite{lin2017feature} as the feature extractor. During the training process, the initial learning rate is set to $0.01$ and the batch size is set to $8$. We train the model for $12$ epochs on the OpenForensics training dataset and the learning rate drops by $1/10$ at the $8$-th and $10$-th epoch. For the FFIW dataset, the model is trained for 6 epochs and the learning rate drops by $1/10$ at the $4$-th and $5$-th epoch. Other settings are the same as on the OpenForensics dataset.
The weight hyper-parameters in Equ.~(\ref{equ:loss}) are set as $\lambda_{1}=0.5$, $\lambda_{2}=0.1$, and $\lambda_{3}=0.1$. For coarse-grained and fine-grained contrastive learning, we set the temperature hyper-parameter as $\tau=0.7$, the exponential hyper-parameter $\beta$ to $0.999$, and the prototypes updating parameter $\alpha$ to $0.9$.  {The coarse-grained learning is conducted at the features of the $3,4,5,6,$ and $7$-th layers, and the weights of these layers are set to $0.1$, $0.2$, $0.4$, $0.7$, and $1.0$.}
For the FFIW dataset, we first utilize YOLOv8 \cite{yolov8_ultralytics} to locate real faces within the dataset. After getting the real faces, we use the OpenCV toolkit dlib \cite{dlib09} to detect the 68 facial landmarks and generate the real facial masks for training and evaluation.

\begin{figure}[t]
\includegraphics[width=\linewidth]{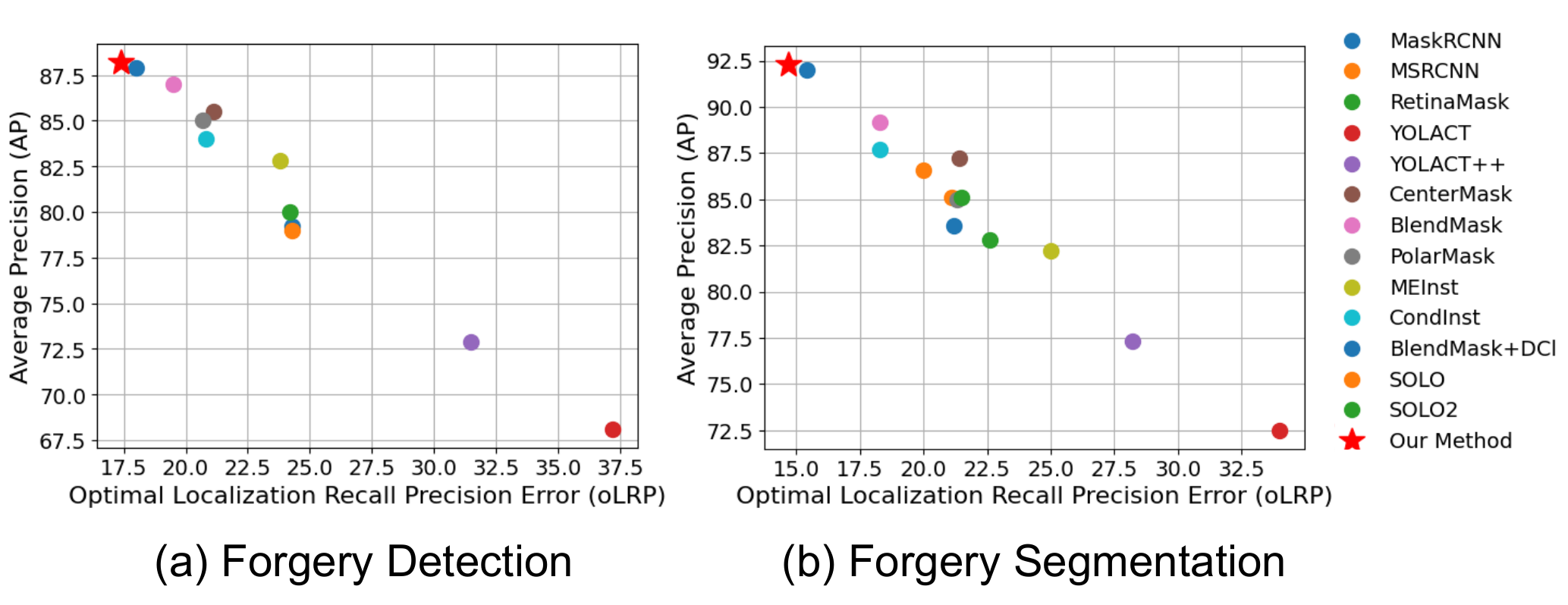} 
 \caption{\small Scatter graph of AP and oLRP compared with other methods on the test-development set. }
  \label{fig:resultdev}
  \vspace{-0.6cm}
\end{figure}

\subsection{Results}
\label{sec:exps-res}
Table~\ref{tab:resultsdev} shows the performance of our method compared with several state-of-the-art methods on the test-development set. The compared methods are adapted from object detectors as in the original OpenForensics experiments \cite{le2021openforensics}. The results show that our method has the best performance for mean AP and oLRP in both detection and segmentation tasks and outperforms the BlendMask baseline by $2.2\%$ at AP and $2.9\%$ at oLRP on average. Similar trends can be observed for other metrics. 
In face forgery detection, the false positive is a critical metric, as it signifies the probability of failure in detecting the forged faces. Table~\ref{tab:resultsdev} demonstrates that the proposed method significantly reduces the false positive rate of oLRP.
For a comprehensive demonstration, we plot the scatter graph of AP and oLRP. Since higher AP and lower oLRP denote better performance, the top-left corner in the graph is the best. As shown in Fig.~\ref{fig:resultdev}, our method is close to the top-left corner, remarkbly outperforming others.
\begin{figure}[t]
\includegraphics[width=\linewidth]{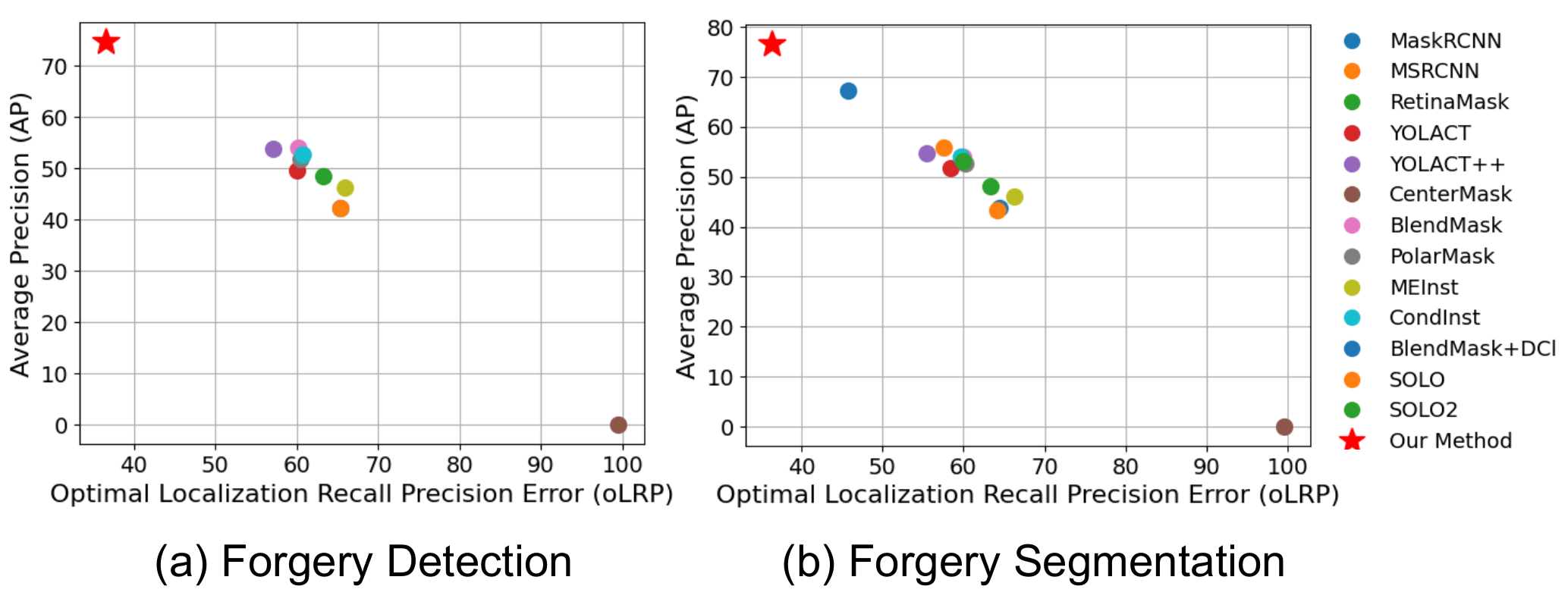} 
 \caption{\small Scatter graph of AP and oLRP compared with other methods on the test-challenge set. }
  \label{fig:resultcha}
  \vspace{-0.5cm}
\end{figure}

\begin{figure*}[!th]
	\centering
	\includegraphics[width=\linewidth]{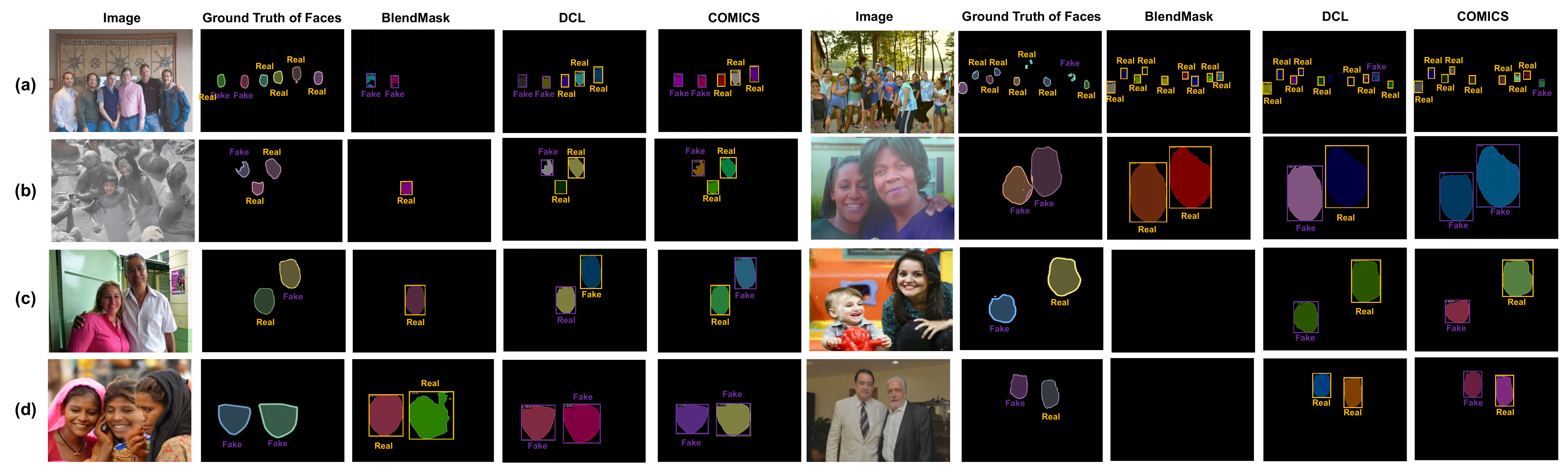} 
        \vspace{-0.7cm}
	\caption{\small  {Visual examples of our method. We use BlendMask \cite{chen2020blendmask}, and DCL \cite{sun2022dual}, as examples for comparison, see Table~\ref{tab:resultsdev} and Table~\ref{tab:resultscha}.}}
	\label{fig:result}
 \vspace{-0.4cm}
\end{figure*}


\begin{table*}[t]
	\centering
	\small
	\caption{\small  Effect of different components of the Contrastive Multi-FaceForensics on OpenForensics dataset.}
        \vspace{-0.3cm}
        Test-development set
        \resizebox{\linewidth}{!}{
  \begin{tabular}{ccc|cccc|cccc|cccc|cccc}
\hline    
\multicolumn{3}{c|}{Method} & \multicolumn{8}{c|}{Multi-Face Forgery Detection}  & \multicolumn{8}{c}{Multi-Face Forgery Segmentation}  \\
 \hline
DA & FEA  & Bi-grained & AP & AP$_{S}$$ $ & AP$_{M} $ & AP$_{L} $ &  oLRP &  oLRP$_{Loc}$ &oLRP$_{FP}$ & oLRP$_{FN} $ & AP  &AP$_{S} $ & AP$_{M}$ & AP$_{L}$ & oLRP & oLRP$_{Loc} $ & oLRP$_{FP} $ & oLRP$_{FN} $ \\
\hline
  -           &    -        &     -        & 87.0 & 32.7 & 86.3 & 88.0 & 19.5 & 6.2 & 2.4 & 6.2 & 89.2 & 19.8 & 87.3 & 91.0 & 18.3 & 5.4 & 2.5 & 6.3 \\
$\checkmark$ &     -        &     -        & 88.4 & 30.9 & 85.6 & 90.1 & 17.4 & 6.2 & 2.7 & 3.4 & 91.7 & 14.8 & 88.3 & 94.1 & 15.2 & 4.9 & 2.7 & 3.7 \\
  -          & $\checkmark$ &     -        & 88.8 & 33.6 & 86.3 & 90.4 & 16.9 & 6.2 & 2.8 & 2.8 & 92.6 & 17.1 & 89.5 & 94.9 & 14.5 & 4.6 & 2.3 & 3.9 \\
  -          &     -        & $\checkmark$ & 88.9 & 31.7 & 86.0 & 90.5 & 17.4 & 6.1 & 3.1 & 3.3 & 92.3 & 15.0 & 88.9 & 94.5 & 15.2 & 4.7 & 3.0 & 3.9 \\ 
 
$\checkmark$ & $\checkmark$ &     -        & 88.2 & 31.5 & 85.4 & 90.0 & 17.4 & 6.3 & 2.7 & 3.2 & 92.2 & 16.4 & 88.9 & 94.6 & 14.8 & 4.8 & 2.9 & 3.3\\
  -          & $\checkmark$ & $\checkmark$ & 88.3 & 32.9 & 85.6 & 90.0 & 17.7 & 6.3 & 2.9 & 3.5 & 91.6 & 16.3 & 88.3 & 93.8 & 15.6 & 5.0 & 2.9 & 3.8 \\
$\checkmark$ &     -        & $\checkmark$ & 88.2 & 31.1 & 85.6 & 89.9 & 17.7 & 6.3 & 2.8 & 3.5 & 91.9 & 14.1 & 88.3 & 94.2 & 15.3 & 4.8 & 2.7 & 3.9 \\
  
$\checkmark$ & $\checkmark$ & $\checkmark$ & 88.2 & 32.3 & 85.7 & 89.9 & 17.4 & 6.2 & 2.5 & 3.5 & 92.3 & 16.3 & 89.1 & 94.6 & 14.7 & 4.7 & 2.6 & 3.6 \\
 \hline
 \label{Tab:IV}
\end{tabular}}
  {Test-challenge} Set\\
\resizebox{\linewidth}{!}{
\begin{tabular}{ccc|cccc|cccc|cccc|cccc}
\hline    
\multicolumn{3}{c|}{Method} & \multicolumn{8}{c|}{Multi-Face Forgery Detection}  & \multicolumn{8}{c}{Multi-Face Forgery Segmentation}  \\
 \hline
DA & FEA  & Bi-grained & AP & AP$_{S}$$ $ & AP$_{M} $ & AP$_{L} $ &  oLRP &  oLRP$_{Loc}$ &oLRP$_{FP}$ & oLRP$_{FN} $ & AP  &AP$_{S} $ & AP$_{M}$ & AP$_{L}$ & oLRP & oLRP$_{Loc} $ & oLRP$_{FP} $ & oLRP$_{FN} $ \\
\hline
  -           &    -        &     -        & 53.9 & 13.5 & 56.6 & 53.5 & 60.2 & 10.6 & 26.5 & 37.4 & 54.0 & 7.1 & 54.5 & 54.5 & 59.9 & 9.8 & 26.4 & 38.4 \\
$\checkmark$ &     -        &     -        & 73.1 & 15.6 & 71.0 & 74.7 & 40.2 & 9.2  & 12.4 & 18.3 & 74.8 & 6.0 & 70.1 & 77.7 & 39.2 & 8.2 & 12.7 & 18.6 \\
  -          & $\checkmark$ &     -        & 67.5 & 16.4 & 66.0 & 68.9 & 47.0 & 9.6  & 16.5 & 24.6 & 68.1 & 6.3 & 62.7 & 71.5 & 46.4 & 8.5 & 16.5 & 25.7 \\
  -          &     -        & $\checkmark$ & 69.0 & 15.3 & 66.0 & 70.9 & 44.9 & 9.6  & 14.9 & 22.5 & 69.4 & 4.9 & 62.6 & 73.1 & 44.3 & 8.6 & 14.3 & 24.1 \\ 
 
$\checkmark$ & $\checkmark$ &     -        & 73.5 & 17.8 & 71.4 & 75.3 & 38.4 & 8.1  & 11.5 & 18.7 & 75.1 & 6.5 & 70.0 & 78.4 & 39.4 & 9.3 & 12.0 & 17.4 \\

  -          & $\checkmark$ & $\checkmark$ & 68.4 & 17.4 & 65.9 & 70.4 & 45.4 & 9.7  & 15.2 &  2.3 & 68.5 & 6.1 & 62.2 & 72.3 & 45.1 & 8.7 & 15.0 & 24.5 \\
$\checkmark$ &     -        & $\checkmark$ & 73.7 & 17.8 & 71.6 & 75.3 & 39.3 & 9.2  & 12.5 & 16.9 & 75.7 & 5.7 & 70.9 & 78.6 & 38.1 & 8.1 & 12.1 & 17.7 \\
  
$\checkmark$ & $\checkmark$ & $\checkmark$ & 74.6 & 19.2 & 72.1 & 76.4 & 36.5 & 9.0  & 10.0 & 16.9 & 76.6 & 7.1 & 71.6 & 79.7 & 36.3 & 7.9 & 9.6  & 17.9 \\
 \hline
\end{tabular}}
\label{tab:abl-all-cha}
 \vspace{-0.3cm}
\end{table*}

The test-challenge set contains more hard images that are processed with different transformations. As shown in Table~\ref{tab:resultscha}, the performance of all methods significantly drops compared to the case in the test-development set. However, compared to other methods, our method has a significant improvement on all metrics. For example, our method outperforms the second-best method BlendMask-DCL by $8.8\%$ at AP and $9.8\%$ at oLRP on average, which greatly corroborates the hypothesis that the proposed bi-grained contrastive learning can capture more effective forgery traces than other adaptation paradigms even in the most challenging scenarios. Similarly, we also plot the scatter graph of AP and oLRP in Fig.~\ref{fig:resultcha}, showing that our method significantly outperforms others in both detection and segmentation tasks.

Compared to the performance on the test-development set, we can observe that our method significantly improves the performance on the test-challenge set. This is because the forgery traces in the test-development set are relatively easy to be captured by all methods. However, the images in the test challenge set are subjected to various transformations, which highly conceal the forgery traces. Unfortunately, traditional methods with direct adaptations are incompetent to deal with such an artifact. In contrast, our method considers the relationship between real and fake faces by coarse- and fine- grained contrastive learning, thus it can capture the discriminative but subtle features for forgery detection.

Fig.~\ref{fig:result} shows several visual examples of the detection results. We use BlendMask as an example for comparison. The left and right group are the results of the test-development set and the test-challenge set, respectively. As shown in Fig.~\ref{fig:result},  BlendMask fails to identify the faces in Fig.~\ref{fig:result}~(f)-left and Fig.~\ref{fig:result}~(d)-right, and misidentifies the fake faces in Fig.~\ref{fig:result}~(e)-left and Fig.~\ref{fig:result}~(f)-right.
In comparison, our proposed COMICS greatly improves the face detection rate. Benefiting from the fine-grained learning, our method can also estimate the boundary of both real and fake faces compared to others, see the right face in Fig.~\ref{fig:result}~(f)-left.

\begin{table*}[t]
	\centering
	\small
	\caption{\small Evaluation on the effect of different components on OpenForensics dataset.}
    \vspace{-0.3cm}
    Test-development set
    \resizebox{\linewidth}{!}{
	\begin{tabular}
    {l|cccc|p{0.9cm}<{\centering}p{0.9cm}<{\centering}p{0.9cm}<{\centering}p{0.9cm}<{\centering}|cccc|p{0.9cm}<{\centering}p{0.9cm}<{\centering}p{0.9cm}<{\centering}p{0.9cm}<{\centering}}
    \hline
    \multirow{2}{*}{Method}    & \multicolumn{8}{c|}{Multi-Face Forgery Detection} & \multicolumn{8}{c}{Multi-Face Forgery Segmentation}  \\
    \cline{2-17} 
    & AP & AP$_{S}$$ $ & AP$_{M} $ & AP$_{L} $ &  oLRP &  oLRP$_{Loc}$ &oLRP$_{FP}$ & oLRP$_{FN} $ & AP  &AP$_{S} $ & AP$_{M}$ & AP$_{L}$ & oLRP & oLRP$_{Loc} $ & oLRP$_{FP} $ & oLRP$_{FN} $ \\
    \hline  
    None          & 87.0 & 32.7 & 86.3 & 88.0 & 19.5 & 6.2 & 2.4 & 6.2 & 89.2 & 19.8 & 87.3 & 91.0 & 18.3 & 5.4 & 2.5 & 6.3 \\
    \hline
    CL            & 88.1 & 32.8 & 85.4 & 89.8 & 17.5 & 6.3 & 2.9 & 3.1 & 92.3 & 16.4 & 89.1 & 94.5 & 14.7 & 4.7 & 2.7 & 3.6\\
    CL w.o. ML    & 86.9 & 31.8 & 85.4 & 87.6 & 22.7 & 6.4 & 4.8 & 8.3 & 90.2 & 14.6 & 89.0 & 92.4 & 21.8 & 4.7 & 5.7 & 8.6\\
    CL 3layer     & 88.0 & 31.5 & 85.4 & 89.2 & 18.1 & 6.3 & 3.0 & 4.3 & 92.3 & 16.7 & 89.1 & 94.5 & 14.8 & 4.7 & 2.8 & 3.7 \\
    CL 2layer     & 88.0 & 31.9 & 85.3 & 89.2 & 19.1 & 6.3 & 2.9 & 5.2 & 92.3 & 17.5 & 89.1 & 94.5 & 14.9 & 4.7 & 2.7 & 3.9 \\
    CL 1layer-7   & 87.9 & 32.0 & 85.3 & 89.0 & 20.2 & 6.4 & 2.9 & 6.2 & 92.2 & 16.7 & 89.1 & 94.4 & 14.8 & 4.7 & 2.7 & 3.8 \\
    CL 1layer-3   & 87.8 & 30.3 & 85.2 & 88.8 & 20.5 & 6.2 & 3.0 & 7.6 & 92.0 & 16.1 & 89.0 & 94.2 & 15.0 & 4.7 & 2.8 & 3.9 \\   
    
    \hline    
    FL            & 88.1 & 31.9 & 85.7 & 89.7 & 17.6 & 6.3 & 3.1 & 3.6 & 92.1 & 16.8 & 88.9 & 94.4 & 14.9 & 4.7 & 2.7 & 3.9 \\
    FL w.o. inter & 87.8 & 32.4 & 85.6 & 89.1 & 17.6 & 6.3 & 3.0 & 3.2 & 92.1 & 16.6 & 89.0 & 94.4 & 15.0 & 4.7 & 2.7 & 3.8 \\
    FL w.o. intra & 88.0 & 31.7 & 85.4 & 89.4 & 17.5 & 6.3 & 2.6 & 3.4 & 92.3 & 16.5 & 89.2 & 94.6 & 14.6 & 4.7 & 2.6 & 3.6 \\
    \hline
    COMICS (Ours) & 88.2 & 32.3 & 85.7 &  89.9& 17.4 & 6.2 & 2.5 & 3.5 & 92.3 & 16.3 & 89.1 & 94.6 & 14.7 & 4.7 & 2.6 & 3.6 \\
     \hline
	\end{tabular}}
Test-challenge set
    \resizebox{\linewidth}{!}{
	\begin{tabular}
    {l|cccc|p{0.9cm}<{\centering}p{0.9cm}<{\centering}p{0.9cm}<{\centering}p{0.9cm}<{\centering}|cccc|p{0.9cm}<{\centering}p{0.9cm}<{\centering}p{0.9cm}<{\centering}p{0.9cm}<{\centering}}
    \hline
    \multirow{2}{*}{Method}    & \multicolumn{8}{c|}{Multi-Face Forgery Detection} & \multicolumn{8}{c}{Multi-Face Forgery Segmentation}  \\
    \cline{2-17} 
    & AP & AP$_{S}$$ $ & AP$_{M} $ & AP$_{L} $ &  oLRP &  oLRP$_{Loc}$ &oLRP$_{FP}$ & oLRP$_{FN} $ & AP  &AP$_{S} $ & AP$_{M}$ & AP$_{L}$ & oLRP & oLRP$_{Loc} $ & oLRP$_{FP} $ & oLRP$_{FN} $ \\
    \hline  
    None       & 53.9 & 13.5 & 56.6 & 53.5 & 60.2 & 10.6 & 26.5 & 37.4 & 54.0 & 7.1 & 54.5 & 54.5 & 59.9 & 9.8 & 26.4 & 38.4 \\
    \hline
    CL            & 74.3 & 18.7 & 72.1 & 76.6 & 37.1 & 9.0 & 9.9  & 16.9 & 76.5 & 6.9 & 71.5 & 79.0 & 36.4 & 7.9 & 9.9 & 17.9\\   
    CL w.o. ML    & 71.8 & 18.4 & 72.2 & 73.6 & 39.7 & 9.0 & 11.5 & 18.7 & 73.8 & 7.1 & 71.5 & 77.8 & 39.8 & 7.9 & 11.4 & 19.3\\ 
    CL 3layer     & 73.6 & 17.5 & 70.3 & 75.7 & 39.1 & 9.1 & 11.8 & 17.6 & 75.5 & 6.6 & 69.1 & 78.0 & 38.0 & 7.9 & 10.9 & 19.0\\
    CL 2layer     & 73.4 & 17.8 & 70.3 & 75.6 & 39.4 & 9.1 & 11.4 & 18.4 & 75.2 & 7.0 & 69.0 & 78.8 & 38.3 & 7.9 & 10.5 & 19.8 \\
    CL 1layer-7   & 72.9 & 17.5 & 70.1 & 75.1 & 40.1 & 9.1 & 11.7 & 19.0 & 74.8 & 6.5 & 68.8 & 78.2 & 38.9 & 7.9 & 11.3 & 20.0\\
    CL 1layer-3   & 71.7 & 17.8 & 71.0 & 74.7 & 39.6 & 9.0 & 11.3 & 18.7 & 74.7 & 7.2 & 68.5 & 78.1 & 38.3 & 7.9 & 10.9 & 19.6 \\ 
   
    \hline    
    FL            & 73.5 & 18.7 & 72.2 & 76.3 & 38.4 & 9.0 & 10.6 & 17.2 & 75.9 & 7.2 & 71.5 & 79.1 & 36.5 & 8.0 & 10.0 & 18.0\\
    FL w.o. inter & 73.6 & 18.7 & 71.8 & 76.5 & 39.6 & 9.0 & 10.4 & 17.7 & 75.6 & 7.0 & 71.5 & 78.7 & 37.4 & 8.0 & 9.9 & 18.6\\
    FL w.o. intra & 73.3 & 19.2 & 72.0 & 76.1 & 38.9 & 9.1 & 9.8 & 17.4 & 75.5 & 6.9 & 71.1 & 78.6 & 37.5 & 8.1 & 9.9 & 18.8\\
    \hline  
    COMICS (Ours)  & 74.6 & 19.2 & 72.1 & 76.4& 36.5 & 9.0 & 10.0 & 16.9 & 76.6 & 7.1 & 71.6 & 79.7& 36.3 & 7.9  & 9.6 & 17.9 \\
    \hline
	\end{tabular}}
	\label{tab:abl-com}
    \vspace{-0.3cm}
    \end{table*}

\begin{table*}[!ht]
    \centering
    \small
    \caption{\small Ablation study on  weights of bi-grained contrastive learning on OpenForensics dataset.}
    \vspace{-0.3cm}
    Test-development set
    \resizebox{\linewidth}{!}{
	\begin{tabular}
    {l|cccc|p{0.9cm}<{\centering}p{0.9cm}<{\centering}p{0.9cm}<{\centering}p{0.9cm}<{\centering}|cccc|p{0.9cm}<{\centering}p{0.9cm}<{\centering}p{0.9cm}<{\centering}p{0.9cm}<{\centering}}
    \hline
    \multirow{2}{*}{$\lambda_{1},\lambda_{2},\lambda_{3}$}    & \multicolumn{8}{c|}{Multi-Face Forgery Detection} & \multicolumn{8}{c}{Multi-Face Forgery Segmentation}  \\
    \cline{2-17} 
    & AP & AP$_{S}$$ $ & AP$_{M} $ & AP$_{L} $ &  oLRP &  oLRP$_{Loc}$ &oLRP$_{FP}$ & oLRP$_{FN} $ & AP  &AP$_{S} $ & AP$_{M}$ & AP$_{L}$ & oLRP & oLRP$_{Loc} $ & oLRP$_{FP} $ & oLRP$_{FN} $ \\
    \hline  
    1.0,1.0,0.1  & 88.1 & 32.3 & 85.4 & 89.7 & 17.7 & 6.3 & 2.7 & 3.6 & 92.1 & 16.5 & 88.8 & 94.4 & 15.0 & 4.7 & 2.8 & 3.7 \\
    1.0,0.5,0.1  & 88.2 & 32.2 & 85.6 & 90.0 & 17.5 & 6.2 & 2.8 & 3.4 & 92.2 & 16.2 & 88.9 & 94.5 & 14.8 & 4.7 & 2.8 & 3.6 \\ 
    1,0.1,0.1.0  & 88.3 & 32.4 & 85.7 & 90.0 & 17.4 & 6.2 & 2.4 & 3.8 & 92.3 & 16.7 & 89.0 & 94.5 & 14.7 & 4.6 & 2.4 & 3.9 \\
    0.5,0.5,1.0  & 88.2 & 32.2 & 85.5 & 89.9 & 17.5 & 6.3 & 2.7 & 3.5 & 92.2 & 16.3 & 89.0 & 94.5 & 14.8 & 4.6 & 2.6 & 3.8 \\
    0.5,0.1,0.1  & 88.2 & 32.3 & 85.7 & 89.9 & 17.4 & 6.2 & 2.5 & 3.5 & 92.3 & 16.3 & 89.1 & 94.6 & 14.7 & 4.7 & 2.6 & 3.6 \\
    \hline
    \end{tabular}}
      Test-challenge set
    \resizebox{\linewidth}{!}{
	\begin{tabular}
    {l|cccc|p{0.9cm}<{\centering}p{0.9cm}<{\centering}p{0.9cm}<{\centering}p{0.9cm}<{\centering}|cccc|p{0.9cm}<{\centering}p{0.9cm}<{\centering}p{0.9cm}<{\centering}p{0.9cm}<{\centering}}
    \hline
    \multirow{2}{*}{$\lambda_{1},\lambda_{2},\lambda_{3}$}    & \multicolumn{8}{c|}{Multi-Face Forgery Detection} & \multicolumn{8}{c}{Multi-Face Forgery Segmentation}  \\
    \cline{2-17} 
    & AP & AP$_{S}$$ $ & AP$_{M} $ & AP$_{L} $ &  oLRP &  oLRP$_{Loc}$ &oLRP$_{FP}$ & oLRP$_{FN} $ & AP  &AP$_{S} $ & AP$_{M}$ & AP$_{L}$ & oLRP & oLRP$_{Loc} $ & oLRP$_{FP} $ & oLRP$_{FN} $ \\
    \hline  
    1.0,1.0,0.1  & 74.8	& 19.2 & 72.2 & 76.6 & 37.1 & 9.0 & 8.9 & 17.3 & 76.7 & 7.1 & 71.4 & 80.0 & 35.8 & 7.8 & 9.1 & 17.7 \\
    1.0,0.5,0.1  & 74.9 & 19.4 & 72.2 & 76.8 & 37.2 & 9.1 & 9.9 & 16.4 & 76.8 & 7.2 & 71.5 & 80.0 & 36.0 & 7.9 & 9.6 & 17.2 \\
    1.0,0.1,0.1  & 74.7 & 19.1 & 72.1 & 76.7 & 37.6 & 9.1 & 10.4 & 16.3 & 76.7 & 7.0 & 71.5 & 79.9 & 36.3 & 7.9 & 10.1 & 17.2 \\
    0.5,0.5,1.0  & 74.7 & 18.8 & 71.9 & 76.6 & 37.3 & 9.1 & 9.7  & 16.6 & 76.8 & 7.0 & 71.5 & 80.0 & 36.0 & 7.9 & 9.3 & 17.5 \\
    0.5,0.1,0.1  & 74.6 & 19.2 & 72.1 & 76.4 & 36.5 & 9.0 & 10.0 & 16.9 & 76.6 & 7.1 & 71.6 & 79.7 & 36.3 & 7.9 & 9.6 & 17.9 \\
      \hline
    \end{tabular}}
    \label{tab:weight}
    \vspace{-0.6cm}
\end{table*}

\subsection{Ablation Study}
\label{ablation}

\noindent{\bf Effect of Different Components.}
This part studies the impact of different components of COMICS, including the proposed data augmentation, frequency enhanced attention, and bi-grained contrastive learning, respectively. The results on test-development and test-challenge sets are shown in Table~\ref{tab:abl-all-cha}. Note that we use BlendMask as the base architecture.  \textit{DA}, \textit{FEA}, and \textit{Bi-grained} denote specially designed data augmentations, frequency enhanced attention module, and the proposed bi-grained contrastive learning, respectively. Without using any component, our method degrades to the base architecture in the first row. 
{Table~\ref{tab:abl-all-cha} shows that the proposed data augmentation brings a great improvement on the test-challenge set by $20.0\%$ at AP and $20.4\%$ on oLRP on average, suggesting the carefully designed data augmentations enhance the ability of forged trace discovery. Compared with base architecture, the frequency enhanced attention and bi-grained contrastive learning separately bring an average increase of $8.6\%$ at AP and $8.7\%$ at oLRP on the test-development and test-challenge set.
Furthermore, we notice the bi-grained contrastive learning improves the AP on small objects, indicating the SRM filter in the frequence enhanced attention concentrates more on the forged traces.}
It can be seen that all the components of our method have a positive impact on the performance of the two sets. Compared with the baseline, our method improves the performance on average by $2.2\%$ at AP and $2.9\%$ at oLRP on test-development set, and $21.7\%$ at AP and $23.2\%$ at oLRP on test-challenge set.




\begin{table*}[t]
    \centering
    \small
    \caption{\small Evaluation on integrating our framework with different architectures on OpenForensics dataset.}
    \vspace{-0.3cm}
    Test-development set
    \resizebox{\linewidth}{!}{
	\begin{tabular}
    {l|cccc|p{0.9cm}<{\centering}p{0.9cm}<{\centering}p{0.9cm}<{\centering}p{0.9cm}<{\centering}|cccc|p{0.9cm}<{\centering}p{0.9cm}<{\centering}p{0.9cm}<{\centering}p{0.9cm}<{\centering}}
    \hline
    \multirow{2}{*}{Method}    & \multicolumn{8}{c|}{Multi-Face Forgery Detection} & \multicolumn{8}{c}{Multi-Face Forgery Segmentation}  \\
    \cline{2-17} 
    & AP & AP$_{S}$$ $ & AP$_{M} $ & AP$_{L} $ &  oLRP &  oLRP$_{Loc}$ &oLRP$_{FP}$ & oLRP$_{FN} $ & AP  &AP$_{S} $ & AP$_{M}$ & AP$_{L}$ & oLRP & oLRP$_{Loc} $ & oLRP$_{FP} $ & oLRP$_{FN} $ \\
     \hline
    MaskRCNN \cite{chen2020blendmask}      & 79.2 & 29.9 & 80.2 & 79.5 & 24.3 & 9.5 & 2.7 & 4.0 & 83.6 & 16.1 & 82.1 & 85.8 & 21.2 & 7.6 & 3.0 & 4.2 \\
     { MaskRCNN+DCL\cite{sun2022dual}}  &  {85.3} &  {28.0} &  {83.9} &  {86.6} &  {21.3} &  {7.3} &  {4.1} &  {4.6} &  {89.1} &  {14.0} &  {86.7} &  {91.4} &  {19.2} &  {5.8} &  {3.5} &  {5.5} \\
    MaskRCNN-COMICS & 85.5 & 26.3 & 83.8 & 86.8 & 21.1 & 7.3 & 3.8 & 4.4 & 89.6 & 13.1 & 86.9 & 91.8 & 19.0 & 5.9 & 3.4 & 5.3 \\
    \hline  
    ConInst \cite{tian2020conditional} & 84.0 & 29.4 & 83.6 & 84.8 & 20.8 & 7.4 & 2.3 & 5.2 & 87.7 & 18.1 & 85.1 & 89.8 & 18.3 & 5.9 & 2.4 & 5.3 \\	
    {ConInst+DCL\cite{sun2022dual}} &  {86.3} &  {28.2} &  {85.0} &  {87.9} &  {19.4} &  {6.1} &  {2.8} & {5.4} &  {89.2} &  {16.6} &  {86.9} &  {92.0} &  {17.3} &  {4.9} &  {2.8} &  {5.6} \\
    ConInst-COMICS  & 87.4 & 29.9 & 85.0 & 89.0 & 18.2 & 6.5 & 3.3 & 3.1 & 91.0 & 14.8 & 87.1 & 93.4 & 16.0 & 5.2 & 2.8 & 4.0 \\   
    \hline
    MEInst \cite{zhang2020mask} & 82.8 & 26.0 & 82.7 & 83.4 & 23.8 & 7.6 & 4.1 & 6.8 & 82.2 & 13.9 & 81.5 & 83.3 & 25.0 & 8.1 & 4.0 & 7.2 \\
     {MEInst+DCL\cite{sun2022dual}} &  {79.6} &  {21.2} &  {76.4} &  {81.5} &  {27.6} &  {9.3} &  {5.1} &  {6.8} &  {80.1} &  {3.6}  &  {76.1} &  {82.5} &  {27.8} &  {9.2} &  {5.3} &  {7.0} \\    
    MEInst-COMICS               & 86.8 & 26.3 & 83.6 & 88.7 & 21.0 & 7.1 & 3.6 & 6.0 & 87.5 & 12.8 & 83.5 & 89.9 & 21.0 & 7.9 & 3.7 & 4.2 \\
    \hline  
    SOLOv2  \cite{wang2020solov2} & - & - & - & - & - & - & - & - & 85.1 & 13.7 & 83.7 & 87.1 & 21.5 & 7.1 & 3.1 & 5.8 \\
     {SOLOv2+DCL\cite{sun2022dual}}  & - & - & - & - & - & - & - & - &  {85.4} &  {11.2} &  {83.9} &  {88.3} &  {21.7} &  {6.7} &  {3.1} &  {6.9} \\
    SOLOv2-COMICS                 & - & - & - & - & - & - & - & - & 85.5 &  7.7 & 82.1 & 88.4 & 21.5 & 7.3 & 3.6 & 5.1 \\
    \hline
     {ST  \cite{liu2021Swin}}  &  {89.2} &  {33.8} &  {88.1} &  {90.3} &  {16.4} &  {5.9} &  {2.9} &  {2.7} &  {92.0} &  {22.0} &  {89.5} &  {94.0} &  {15.2} &  {5.1} &  {3.0} &  {2.9} \\
    {ST+DCL \cite{sun2022dual}}&  {82.0} &  {27.7} &  {82.9} &  {82.5} &  {23.2} &  {8.5} &  {3.5} &  {4.6} &  {87.0} &  {15.8} &  {85.3} &  {89.0} &  {20.1} &  {6.6} &  {3.7} &  {4.8} \\
     {ST+COMICS}               &  {88.9} &  {31.0} &  {87.9} &  {90.0} &  {16.6} &  {6.1} &  {2.8} &  {3.0} &  {91.2} &  {20.4} &  {88.5} &  {93.2} &  {16.1} &  {5.3} &  {2.9} &  {3.7} \\
    \hline
    \end{tabular}}
    Test-challenge set
    \resizebox{\linewidth}{!}{
	\begin{tabular}
    {l|cccc|p{0.9cm}<{\centering}p{0.9cm}<{\centering}p{0.9cm}<{\centering}p{0.9cm}<{\centering}|cccc|p{0.9cm}<{\centering}p{0.9cm}<{\centering}p{0.9cm}<{\centering}p{0.9cm}<{\centering}}
    \hline
    \multirow{2}{*}{Method}    & \multicolumn{8}{c|}{Multi-Face Forgery Detection} & \multicolumn{8}{c}{Multi-Face Forgery Segmentation}  \\
    \cline{2-17} 
    & AP & AP$_{S}$$ $ & AP$_{M} $ & AP$_{L} $ &  oLRP &  oLRP$_{Loc}$ &oLRP$_{FP}$ & oLRP$_{FN} $ & AP  &AP$_{S} $ & AP$_{M}$ & AP$_{L}$ & oLRP & oLRP$_{Loc} $ & oLRP$_{FP} $ & oLRP$_{FN} $ \\
     \hline
    MaskRCNN \cite{chen2020blendmask} & 42.1 & 11.8 & 46.2 & 40.5 & 65.4 & 13.6 & 29.3 & 40.0 & 43.7 & 4.7 & 44.3 & 44.0 & 64.4 & 11.8 & 29.4 & 41.2 \\
     {MaskRCNN+DCL\cite{sun2022dual}}  &  {56.0} &  {13.1} &  {58.0} &  {55.4} &  {54.4} &  {10.7} &  {21.4} &  {30.3} &  {57.4} &  {4.6} &  {56.0} &  {58.4} &  {53.8} &  {9.4} &  {20.6} &  {31.8}  \\
    MaskRCNN-COMICS & 65.8 & 14.2 & 65.1 & 66.5 & 46.1 & 10.2 & 15.5 & 22.7 & 68.3 & 5.3 & 65.3 & 70.2 & 44.8 & 8.8 & 15.4 & 23.3 \\
    \hline  
    ConInst \cite{tian2020conditional} & 52.7 & 12.6 & 55.3 & 51.8 & 60.7 & 11.5 & 28.3 & 35.3 & 54.1 & 6.5 & 55.2 & 53.8 & 59.6 & 10.0 & 26.7 & 37.3 \\  
     {ConInst+DCL\cite{sun2022dual}} &  {67.6} &  {13.5} &  {65.0} &  {69.6} &  {45.6} & { 9.2} & { 14.9} &  {24.8} &  {69.2} &  {5.1} &  {62.9} & { 72.5} &  {44.8 }&  {8.3} &  {15.1} &  {25.0} \\
    ConInst-COMICS  & 72.1 & 17.4 & 70.6 & 73.5 & 40.7 & 9.0  & 11.7 & 20.3 & 74.4 & 6.5 & 70.5 & 76.9 & 39.5 & 8.0  & 11.8 & 20.4 \\   
    \hline
    MEInst  \cite{zhang2020mask} & 46.1 & 8.6  & 49.9 & 44.9 & 65.9 & 12.4 & 34.6 & 39.7 & 46.0 & 3.8 & 49.0 & 45.2 & 66.2 & 12.6 & 34.8 & 39.8 \\
     {MEInst+DCL\cite{sun2022dual}} &  {62.1} &  {14.9} &  {59.9} &  {63.3} &  {50.9} &  {11.9} &  {19.4} &  {23.5} &  {62.6} &  {1.9} &  {57.7} &  {64.9} &  {51.0} &  {11.7} &  {18.5} &  {24.9}\\
    MEInst-COMICS                & 67.2 & 15.8 & 63.8 & 69.0 & 44.3 & 11.7 & 12.3 & 19.1 & 68.0 & 1.9 & 63.6 & 70.8 & 44.2 & 11.4 & 12.2 & 19.5 \\
    \hline
    SOLOv2  \cite{wang2020solov2}& - & - & - & - & - & - & - & - & 53.2 & 3.6 & 52.1 & 54.0 & 59.6 & 11.0 & 24.5 & 37.2 \\
     {SOLOv2+DCL\cite{sun2022dual}} & - & - & - & - & - & - & - & - &  {65.4} &  {3.2} &  {60.3} &  {68.2} &  {49.5} &  {11.0} &  {17.7} &  {24.6} \\
    SOLOv2-COMICS & - & - & - & - & - & - & - & - & 70.2 & 2.9 & 65.3 & 73.0 & 43.2 & 11.0 & 11.9 & 19.2 \\
    \hline
        {ST \cite{liu2021Swin}}    &  {52.4} &  {14.6} &  {53.1} &  {52.0} &  {56.6} &  {8.9}  &  {23.0} &  {35.0} &  {53.6} &  {7.6} &  {52.3} &  {53.9} &  {56.3} &  {7.8}  &  {22.1} &  {36.4} \\
        {ST+DCL \cite{sun2022dual}}&  {53.5} &  {13.5} &  {59.4} &  {52.2} &  {58.1} &  {12.1} &  {23.8} &  {32.4} &  {56.4} &  {6.5} &  {60.2} &  {56.0} & {56.6} &  {10.6} &  {24.3} &  {32.2} \\
        {ST+COMICS}                &  {57.5} &  {14.5} &  {58.4} &  {57.0} &  {52.8} &  {9.2}  &  {20.9} &  {30.7} &  {58.8} &  {7.3} &  {57.4} &  {59.4} &  {52.3} &  {8.2}  &  {21.0} &  {31.6} \\
       \hline
    \end{tabular}}
	\label{tab:abl-model}
    \vspace{-0.3cm}
    \end{table*}

\begin{table*}[t]
    \centering
    \small
    \caption{\small {Comparisons of our method on the FFIW dataset.}}
    \vspace{-0.3cm}
    \resizebox{\linewidth}{!}{
	\begin{tabular}
    {l|cccc|p{0.9cm}<{\centering}p{0.9cm}<{\centering}p{0.9cm}<{\centering}p{0.9cm}<{\centering}|cccc|p{0.9cm}<{\centering}p{0.9cm}<{\centering}p{0.9cm}<{\centering}p{0.9cm}<{\centering}}
    \hline
    \multirow{2}{*}{Method}    & \multicolumn{8}{c|}{Multi-Face Forgery Detection} & \multicolumn{8}{c}{Multi-Face Forgery Segmentation}  \\
    \cline{2-17} 
    & AP & AP$_{S}$$ $ & AP$_{M} $ & AP$_{L} $ &  oLRP &  oLRP$_{Loc}$ &oLRP$_{FP}$ & oLRP$_{FN} $ & AP  &AP$_{S} $ & AP$_{M}$ & AP$_{L}$ & oLRP & oLRP$_{Loc} $ & oLRP$_{FP} $ & oLRP$_{FN} $ \\
    \hline  
   BlendMask  \cite{chen2020blendmask}    & 69.7 & 23.4 & 54.6 & 78.3 & 34.2 & 11.2 & 7.2 & 9.6  & 73.7 & 19.9 & 60.9 & 81.2 & 34.8 & 5.7 & 13.1 & 17.0 \\
    {BlendMask+DCL\cite{sun2022dual}} &  {70.4} &  {23.3 }&  {55.7} &  {79.9} &  {33.7} &  {10.4} &  {7.0} &  {10.8} &  {74.0} &  {21.3} &  {59.2} &  {80.9} &  {35.0} &  {5.8} &  {13.5} &  {17.0}\\
   BlendMask+COMICS  & 71.3 & 20.7 & 56.6 & 79.7 & 33.2 & 10.6 & 6.5 & 10.4 & 75.1 & 20.5 & 60.6 & 81.8 & 34.7 & 5.6 & 12.8 & 17.3\\
    \hline
   MaskRCNN  \cite{he2017mask} & 54.1 & 12.7 & 38.5 & 63.1 & 52.3 & 10.6 & 12.8 & 32.2 & 50.9 & 15.0 & 41.1 & 55.5 & 55.5 & 8.6 & 19.9 & 37.1 \\
    {MaskRCNN+DCL\cite{sun2022dual}} &   {55.1} &   {13.4} &   {42.4 }&   {62.9} &   {51.3} &  {10.7} &  {14.9} &  {32.7} &  {52.7} &  {17.5} &  {42.2} &  {56.7} &  {54.8} &  {7.8} &  {18.7} &  {36.3}\\
   MaskRCNN+COMICS                       & 56.5 & 12.6 & 44.5 & 63.8 & 51.1 &  9.5 & 14.3 & 32.3 & 52.8 & 17.0 & 40.8 & 57.4 & 54.3 & 7.8 & 18.5 & 37.2 \\

   \hline
   ConInst  \cite{tian2020conditional}  & 70.3 & 18.3 & 55.1 & 79.6 & 33.2 & 10.6 & 7.6 & 9.4 & 73.9 & 18.5 & 58.7 & 81.6 & 35.4 & 6.0 & 13.2 & 17.1 \\
    {ConInst+DCL\cite{sun2022dual}} &  {69.8} &  {22.0} &  {55.5} &  {78.3} &  {33.9} &  {10.7} &  {8.1} &  {9.5} & {73.3} &  {18.5} &  {59.2} &  {80.8} &  {35.7} &  {6.0} &  {13.6} &  {17.3} \\
   ConInst+COMICS                       & 71.3 & 20.2 & 54.1 & 80.6 & 33.0 & 10.6 & 6.6 & 9.8 & 73.9 & 19.2 & 58.0 & 81.0 & 34.9 & 6.0 & 12.1 & 17.4 \\
   \hline
   MEInst  \cite{zhang2020mask}         & 64.2 & 25.0 & 48.1 & 73.0 & 46.8 & 7.4  & 20.6 & 25.3 & 66.1 & 12.2 & 51.5 & 72.6 & 46.9 & 7.4 & 21.5 & 24.6\\
    {MEInst+DCL\cite{sun2022dual}}  &  {64.4} &  {25.6} & { 48.0} &  {73.0} &  {46.7} &  {11.2 }&  {17.2 }&  {19.5} &  {66.1} &  {12.6} &  {51.8} &  {72.5 }&  {46.9} & { 7.7 } & { 16.8 }&  {19.5} \\
   MEInst+COMICS                        & 64.5 & 25.1 & 48.3 & 73.4 & 46.2 & 11.2 & 16.9 & 18.7 & 66.2 & 12.1 & 51.7 & 72.7 & 46.2 & 11.3 & 16.8 & 18.6 \\
     \hline
   SOLOv2 \cite{wang2020solov2}       & - & - & - & - & - & - & - & - & 67.7 & 7.5 & 49.8 & 75.9 & 44.4 & 7.4 & 17.5 & 24.2 \\
  {SOLOv2+DCL\cite{sun2022dual}}& - & - & - & - & - & - & - & - &  {62.5} &  {7.7} &  {44.2} &  {70.5} &  {51.5} & { 7.5} &  {24.9} &  {29.4}\\
   SOLOv2+COMICS               & - & - & - & - & - & - & - & - & 67.8 & 7.4 & 50.0 & 76.1 & 44.4 & 7.4  & 17.4 & 24.4 \\
    \hline
     {ST \cite{liu2021Swin}}    &  {86.4} &  {56.5} &  {78.4} &  {89.9}&  {26.3} &  {5.4} &  {8.3} &  {10.4} &  {67.3} &  {22.1} &  {55.0}&  {73.6} &  {40.2} &  {8.5} &  {14.0}&  {18.2} \\
        {ST+DCL \cite{sun2022dual}} &  {84.8} &  {60.2} &  {76.3} &  {88.4}&  {28.7} &  {6.1} &  {9.1}&  {11.6} &  {66.7} &  {24.1} &  {54.4}&  {73.0} &  {41.2} &  {8.8} &  {14.7}  &  {18.4}\\       
        {ST+COMICS}                &  {87.4} &  {57.1} &  {78.8} &  {91.0}&  {25.5} &  {5.4} &  {7.6} &  {10.5}&  {67.6} &  {23.0} &  {56.8} &  {73.5}&  {39.7} &  {8.3} &  {14.3} &  {17.7}   \\
      
       \hline    \end{tabular}}
    \label{tab:ffiw}
    \vspace{-0.3cm}
\end{table*}

\noindent{\bf Coarse-grained Contrastive Learning.}
This part studies the effect of coarse-grained contrastive learning and the multi-layer learning as described in Equ.~\ref{eqn:multiscale} on both the test-development set and the test-challenge set. Table~\ref{tab:abl-com} shows the performance with and without coarse-grained contrastive learning. Note that {\em None} denotes no bi-grained contrastive learning is used, and {\em CL} denotes coarse-grained contrastive learning is used.
The results reveal that coarse-grained contrastive learning improves the performance, especially on large objects, and boosts the performance on both the test-development set and the test-challenge set by $2.1\%$ at AP, $2.8\%$ at oLRP and $21.5\%$ at AP, $23.3\%$ at oLRP respectively. 
To demonstrate the effect of multi-layer aggregation, we remove it from coarse-grained learning by fusing the proposals from all layers as a whole, denoted as {\em CL w.o. ML}. As shown in Table~\ref{tab:abl-com} ({\em CL w.o. ML}), the performance drops by $1.7\%$ at AP, $6.2\%$ at oLRP on test-development set and $2.6\%$ at AP, $3.0\%$ at oLRP on test-challenge set, which demonstrates the effect of multi-layer.
Furthermore, we evaluate the effect of layers used for coarse contrastive learning. We select the $3$-rd and $7$-th layer, denoting them as \textit{CL 1layer-3} and \textit{CL 1layer-7} respectively. Additionally, we explore configurations with three and five layers for coarse contrastive learning: the $6$ and $7$-th layers for \textit{CL 2layer}, and the $5,~6$ and $7$-th layers for \textit{CL 3layer}. Note that the full setting \textit{CL} utilizes all the $3$-rd$ \sim 7$-th layers. As shown in Table~\ref{tab:abl-com},
employing coarse contrastive learning across all five layers yields the best performance.

\smallskip
\noindent{\bf Fine-grained Contrastive Learning.}
This part studies the effect of using fine-grained contrastive learning ({\em FL}), only using intra-face loss ({\em FL w.o. inter}) and only using inter-face loss ({\em FL w.o. intra}). The results are also shown in Table~\ref{tab:abl-com}. By adding the fine-grained contrastive leaning, the performance is improved by $2.0\%$ at AP and $2.7\%$ at oLRP on the test-development set, and  $20.8\%$ at AP and $22.6\%$ at oLRP on the test-challenge set. Both the intra-face and inter-face loss contribute to improvement of the performance.


\smallskip
\noindent{\bf Different Weights in Loss.}
Table~\ref{tab:weight} shows the performance of different weights in the loss of bi-grained contrastive learning on the test-development set (top) and test-challenge set (bottom), respectively. Specifically, $\lambda_{1}$ is selected from $\{1, 0.5\}$, $\lambda_{2}$ is selected from $\{1, 0.5, 0.1\}$, $\lambda_{3}$ is selected from $\{1, 0.1\}$. The results show that different weights achieve similar performance, which demonstrates the robustness of our method against different weights. We use $\lambda_{1}=0.5$, $\lambda_{2}=0.1$, and $\lambda_{3}=0.1$ as the standard setting of our method. 


\smallskip
\noindent{\bf Combining COMICS with Different Architectures.}
To demonstrate the effectiveness of the proposed bi-grained contrastive learning, we integrate it into other different detection architectures, {\it i.e.}, MaskRCNN \cite{he2017mask}, ConInst \cite{tian2020conditional}, MEInst \cite{zhang2020mask}, SOLOv2 \cite{wang2020solov2},  {and Swin Transformer \cite{liu2021Swin} (denoted as \textit{ST}). Note that the Swin Transformer is a more recent strong Transformer-based object detector. We then compare the performance of COMICS with both the baseline and the implementation of DCL.} The results in Table~\ref{tab:abl-model} show that the proposed framework improves the detection and segmentation performance on all the benchmarks. 
 {Compared to DCL, COMICS achieves an average improvement of $3.2\%$ at AP and $3.0\%$ at oLRP on the test-development set, and $5.8\%$ at AP and $6.3\%$ at oLRP on the test-challenge set.}




\smallskip
\noindent{\bf {Evaluation on FFIW Dataset.}}
The FFIW dataset \cite{Zhou_2021_CVPR} only provides ground truth for fake faces in DeepFake videos. For evaluation, we utilize YOLOv8 for face detection and dlib to generate the real face mask.
In Table~\ref{tab:ffiw}, we evaluate the performance of the proposed COMICS framework on BelandMask, ConInst, MaskRCNN and SOLOv2. COMICS improves the performance on all the benchmarks.
 {Compared to DCL, the performance is improved by $1.6\%$ at AP and $1.5\%$ at oLRP on average.} It is worth noting that the performance improvement on the FFIW dataset is not as significant as observed on the OpenForensics dataset. One possible explanation could be that the inaccuracies of the self-generating real face masks constrain the overall performance on the FFIW dataset.

\smallskip
 \noindent{ {\bf Evaluation on AI-Generated Images.}}  {To evaluate performance against AI-generated images, we randomly select 1,000 images from the OpenForensics test-dev dataset and use BLIP-2 \cite{li2023blip} to generate image captions as prompts for Stable Diffusion  v1.5 \cite{Rombach_2022_CVPR}. Additionally, we evaluated COMICS on the image subsets of Diff \cite{cheng2024diffusion} generated by Stable Diffusion XL (SDXL) \cite{podell2023sdxl}, Midjourney \cite{midjourney}, and DreamBooth \cite{ruiz2023dreambooth}. We report the hit rate (the rate at which at least one forged face is detected), the miss rate (the rate at which no faces are detected), and the error rate (the rate at which faces are mistakenly classified as real) for evaluation. Note that the evaluated COMICS is trained on the OpenForensics dataset without finetuning on AI-generated images. The fake faces in the OpenForensics dataset are created through face swapping, leading COMICS to focus on localized forgery artifacts. In contrast, AI-generated images exhibit globally distributed forgery traces, making the detection of AI-generated faces more challenging due to the differences in textual artifacts. As indicated in Table~\ref{tab:aiimg}, ST-COMICS outperforms BlendMask-COMICS on miss rate  due to the strong detection capability of its object detector. However, BlendMask-COMICS achieves a zero error rate, indicating that it correctly classifies all detected faces. We believe that this part of experiment can provide some empirical guidance to the development of face forgery detection to cope the potential risk brought by quick progress in AIGC domain.}
\begin{table*}[ht]
\centering
\small
 \caption{\small  Performance of COMICS on the AI-generated images.} 
 \vspace{-0.3cm}
\label{tab:aiimg}
\begin{tabular}{c|c|ccc|ccc}
\hline
\multirow{2}{*}{Method}  & \multirow{2}{*}{$\#$ of images}  &\multicolumn{3}{c|}{BlendMask+COMICS} & \multicolumn{3}{c}{Swin Transformer+COMICS }  \\
\cline{3-8}
  &  & Miss Rate$\downarrow$  & Hit Rate$\uparrow$  & Error Rate$\downarrow$ & Miss Rate$\downarrow$ & Hit Rate$\uparrow$  & Error Rate$\downarrow$\\	
\hline  
SD  v1.5   & 1000  & 0.25 & 0.75 & 0 & 0.06 & 0.64 & 0.30 \\
SD XL      & 5508 & 0.81 & 0.19 & 0 & 0.24 & 0.31 & 0.45 \\
Midjourney & 5647 & 0.76 & 0.24 & 0 & 0.18 & 0.29 & 0.53 \\
DreamBooth & 3913 & 0.98 & 0.02 & 0 & 0.15 & 0.54 & 0.32 \\
\hline
\end{tabular}
\vspace{-0.6cm}
\end{table*}

\smallskip
\noindent{\bf Limitations.}
{COMICS is designed on recent anchor-based or anchor-free detection architectures. Despite that it demonstrates a certain level of improvement on small faces as shown in Table~\ref{Tab:IV}, it may still inherit the drawbacks of such architecture of deal with small objects. This limitation is demonstrated by the results in Table~\ref{tab:resultsdev}, where our method drops slightly in detecting small forgery faces.}
 
\section{Conclusion}
We introduce COMICS for multi-face forgery detection. Different from other approaches, our method can simultaneously discover multiple real and fake faces in a glimpse, using the newly proposed bi-grained contrastive learning. Compared to the existing adaption strategy from object detectors, the bi-grained contrastive learning is devoted to capturing the forgery traces in single-stage architecture from the perspectives of coarse-grained and fine-grained. The coarse-grained learning focuses on the proposal-wise discrepancy in a multi-layer manner, guided by the proposal generator. The fine-grained learning focuses on the feature maps obtained by mask predictor and targets the pixel-wise discrepancy in both inter-face and intra-face configurations. Extensive experiments are conducted on the OpenForensics and FFIW datasets, showing that our method significantly outperforms other methods in challenging scenarios. We also conduct comprehensive ablation studies to investigate the effect of different components and settings in the bi-grained contrastive learning.


In future study, we will further explore extracting more discriminative features with contrastive learning and reducing the detection rate gap between large and small faces in multi-face forgery detection.
We also aim to extend the COMICS framework to broader application in other vision tasks, due to its fine-grained feature learning ability.


\bibliographystyle{IEEEtran}
\bibliography{ref}

\begin{IEEEbiography}[{\includegraphics[width=1in,height=1.25in,clip,keepaspectratio]{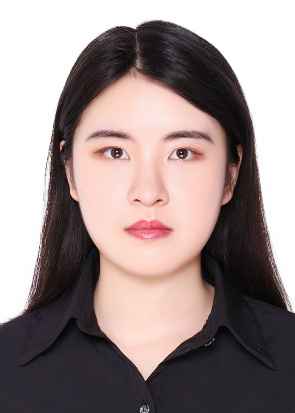}}]{Cong Zhang} (Member, IEEE) received B.S. degree from University of Science and Technology Beijing in 2017 and M.S. degree  from University of Chinese Academy of Sciences in 2020, respectively. She is currently pursuing the PhD degree with the School of Computer Science and Technology, University of Chinese Academy of Sciences. Her research interests include  multimedia forensics and adversarial robustness.\end{IEEEbiography}
\begin{IEEEbiography}[{\includegraphics[width=1in,height=1.25in,clip,keepaspectratio]{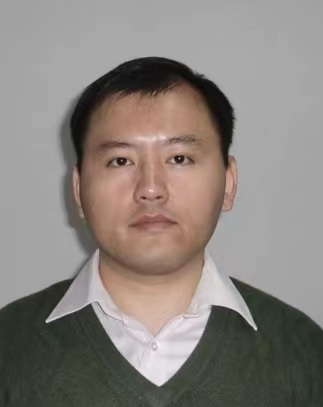}}]{Honggang Qi} (Member, IEEE) received the M.S. degree in computer science from Northeast University, Shenyang, China, in 2002, and the Ph.D. degree in computer science from the Institute of Computing Technology, Chinese Academy of Sciences, Beijing, China, in 2008. He is currently a Professor with the School of Computer Science and Technology, University of Chinese Academy of Sciences. His current research interests include computer vision, video coding, and very large scale integration design. \end{IEEEbiography}
\begin{IEEEbiography}[{\includegraphics[width=1in,height=1.25in,clip,keepaspectratio]{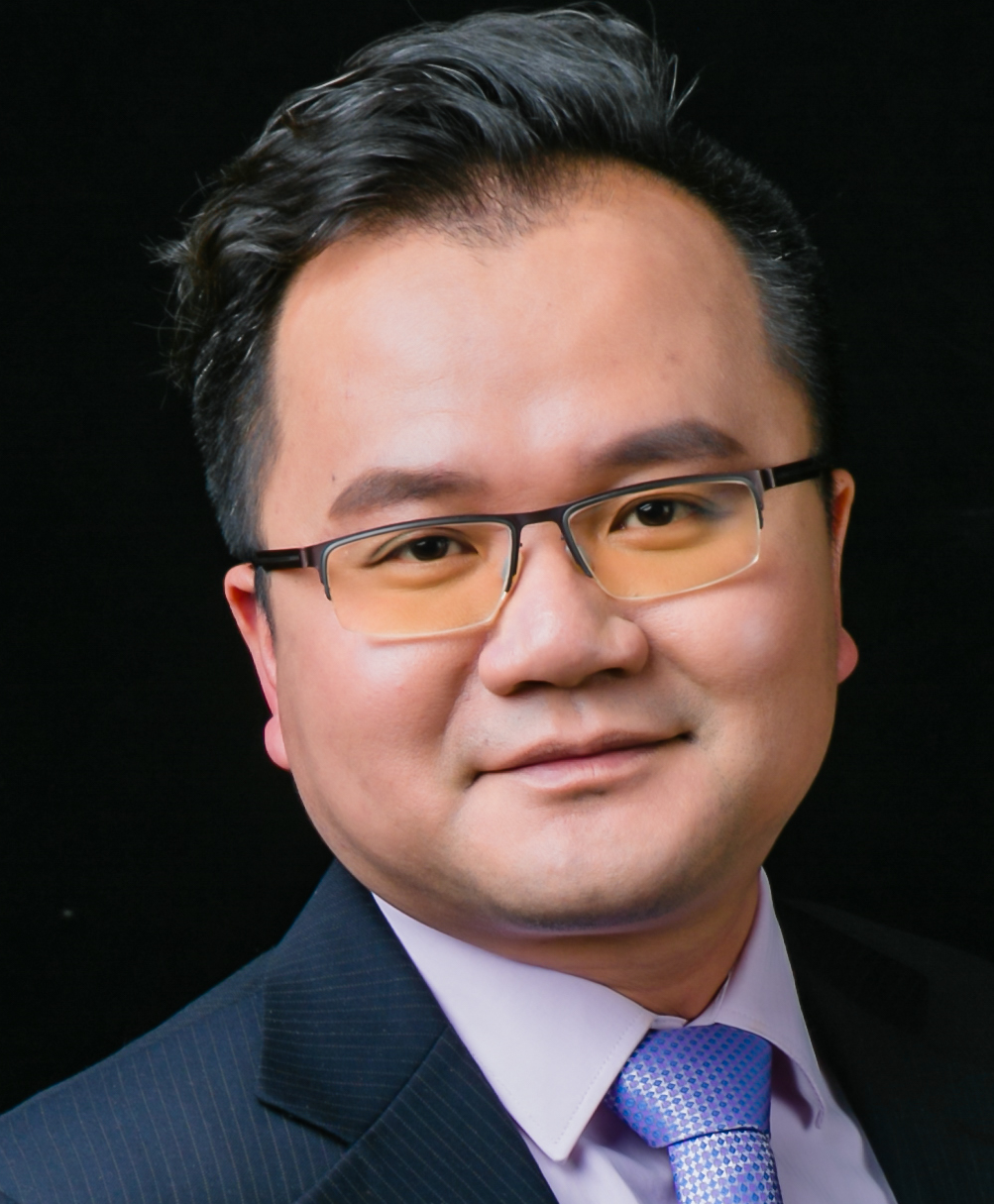}}]{Shuhui Wang} (Member, IEEE) received the B.S. degree in electronics engineering from Tsinghua University, Beijing, China, in 2006, and the Ph.D. degree from the Institute of Computing Technology, Chinese Academy of Sciences, Beijing, China, in 2012. He is currently a Full Professor with the Institute of Computing Technology, Chinese Academy of Sciences. He is also with the Key Laboratory of Intelligent Information Processing, Chinese Academy of Sciences. His research interests include image/video understanding/retrieval, cross-media analysis, visual-textual knowledge extraction and neural computational theories.
\end{IEEEbiography}
\begin{IEEEbiography}[{\includegraphics[width=1in,height=1.25in,clip,keepaspectratio]{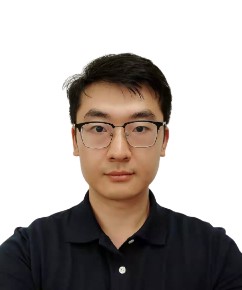}}]{Yuezun Li} (Member, IEEE) is a lecturer in the Center on Artificial Intelligence, at Ocean University of China. He was a Senior Research Scientist at the Department of Computer Science and Engineering of the University at Buffalo, SUNY. He received Ph.D. degree in computer science at University at Albany, SUNY in 2020. He received M.S. degree in Computer Science in 2015 and B.S. degree in Software Engineering in 2012 at Shandong University. Dr. Li’s research interest is mainly focused on artificial intelligence security and multimedia forensics. His work has been published in peer-reviewed conferences and journals, including ICCV, CVPR, TCSVT, TNNL, etc. \end{IEEEbiography}
\begin{IEEEbiography}[{\includegraphics[width=1in,height=1.25in,clip,keepaspectratio]{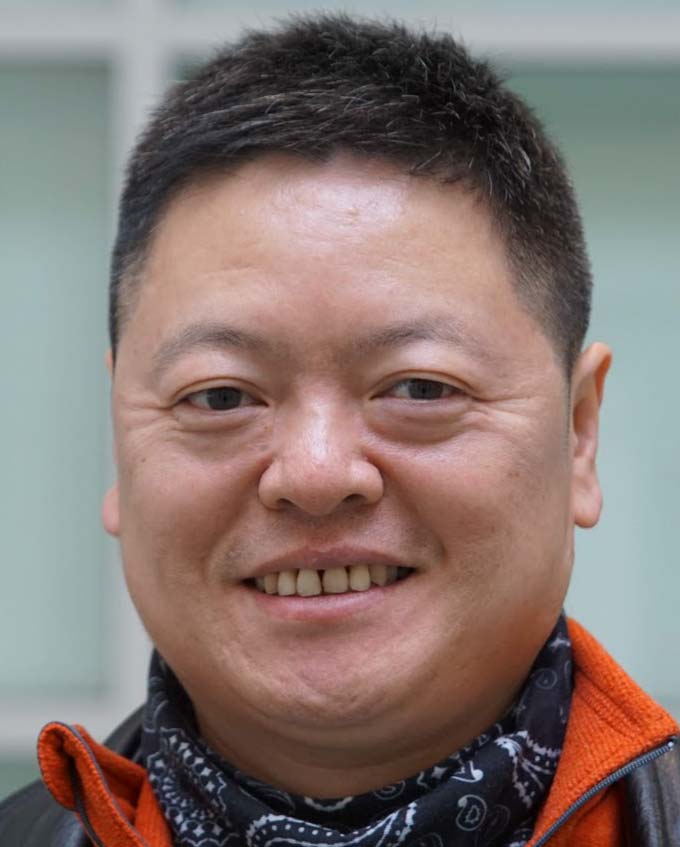}}]{Siwei Lyu} (Fellow, IEEE) received the B.S. degree in information science and the M.S. degree in computer science from Peking University, China, in 1997 and 2000, respectively, and the Ph.D. degree in computer science from Dartmouth College in 2005. From 2005 to 2008, he was a Post-Doctoral Research Associate with Howard Hughes Medical Institute and the Center for Neural Science of New York University. He is currently a Full Professor in computer science with the University at Buffalo (State University of New York at Buffalo). His research interests include digital media forensics, computer vision, and machine learning \end{IEEEbiography}

\end{document}